\documentclass[11pt,a4paper]{article}
\pdfoutput=1
\usepackage{acl_general}
\usepackage{times}
\usepackage{latexsym}
\usepackage[T1]{fontenc}
\usepackage[utf8]{inputenc}
\usepackage{bm}
\usepackage{lipsum}
\usepackage{float}
\usepackage{tikz-dependency}
\usepackage{tikz}
\usepackage{wrapfig}
\usepackage{natbib}
\usepackage{amsmath}
\usepackage{colortbl}
\definecolor{LightRed}{rgb}{1,0.9,0.9}
\usepackage{algorithm2e}
\usepackage{enumitem}
\usepackage{booktabs}
\usepackage{adjustbox}
\usepackage{rotating}
\usepackage{graphicx}
\usepackage{subcaption}
\usepackage{fixmetodonotes}
\usepackage{tabularx}
\usepackage{amssymb}

\usepackage{microtype}

\newcommand\blfootnote[1]{%
  \begingroup
  \renewcommand\thefootnote{}\footnote{#1}%
  \addtocounter{footnote}{-1}%
  \endgroup
}

\newcommand{\MAMLPrime}{\textsc{maml-}}

\title{Meta-Learning for Fast Cross-Lingual Adaptation in Dependency Parsing}

\author{\textbf{Anna Langedijk$^{\bigstar,1}$}, Verna Dankers$^{1,2}$, Phillip Lippe$^1$, Sander Bos$^1$, \\ \textbf{Bryan Cardenas Guevara}$^1${\bf,} \textbf{Helen Yannakoudakis}$^3${\bf, and} \textbf{Ekaterina Shutova}$^1$ \\
\normalfont \normalsize{$^1$Institute for Logic, Language and Computation, University of Amsterdam}\\
\normalsize{$^2$Institute of Language, Cognition and Computation, University of Edinburgh}\\
\normalsize{$^3$Department of Informatics, King’s College London}}

\begin{document}
\maketitle
\blfootnote{$^\bigstar$ Corresponding author: \href{annalangedijk@gmail.com}{annalangedijk@gmail.com}.}

\begin{abstract}
Meta-learning, or learning to learn, is a technique that can help to overcome resource scarcity in cross-lingual NLP problems, by enabling fast adaptation to new tasks. 
We apply model-agnostic meta-learning (\textsc{maml}) to the task of cross-lingual dependency parsing. We train our model on a diverse set of languages to learn a parameter initialization that can adapt quickly to new languages.
We find that meta-learning with pre-training can significantly improve upon the performance of language transfer and standard supervised learning baselines for a variety of unseen, typologically diverse, and low-resource languages, in a few-shot learning setup. 

\end{abstract}

\section{Introduction}
\label{section:intro}
The field of natural language processing (NLP) has seen substantial performance improvements due to large-scale language model pre-training \cite{devlin-etal-2019-bert}. Whilst providing an informed starting point for subsequent task-specific fine-tuning, such models still require large annotated training sets for the task at hand \cite{yogatama}. This limits their applicability to a handful of languages for which such resources are available and leads to an imbalance in NLP technology's quality and availability across linguistic communities. Aiming to address this problem, recent research has focused on the development of multilingual sentence encoders, such as multilingual BERT (mBERT) \cite{devlin-etal-2019-bert} and XLM-R \cite{conneau2019unsupervised}, trained on as many as 93 languages. Such pre-trained multilingual encoders enable zero-shot transfer of task-specific models across languages \cite{wu2019}, offering a possible solution to resource scarcity.
Zero-shot transfer, however, is most successful among typologically similar, high-resource languages, and less so for languages distant from the training languages and in resource-lean scenarios \cite{lauscher2020zero}. This stresses the need to develop techniques for fast cross-lingual model adaptation, that can transfer knowledge across a wide range of typologically diverse languages with limited supervision.

In this paper, we focus on the task of universal dependency (UD) parsing and present a novel approach for effective and resource-lean cross-lingual parser adaptation via meta-learning, requiring only a small number of training examples per language (which are easy to obtain even for low-resource languages).
Meta-learning is a learning paradigm that leverages previous experience from a set of tasks to solve a new task efficiently. As our goal is fast cross-lingual model adaptation, we focus on optimization-based meta-learning, where the main objective is to find a set of initial parameters from which rapid adaption to a variety of different tasks becomes possible \citep{hospedales2020metalearning}. Optimization-based meta-learning has been successfully applied to a variety of NLP tasks. Notable examples include neural machine translation~\cite{gu2018meta}, semantic parsing~\cite{huang-etal-2018-natural}, pre-training text
representations~\cite{lv2020pretraining}, word sense disambiguation~\cite{shutova2020} and cross-lingual natural language inference and question answering~\cite{nooralahzadeh2020}. To the best of our knowledge, meta-learning has not yet been explored in the context of dependency parsing.

We take inspiration from recent research on universal dependency parsing \cite{tran2019,kondratyuk2019}. We employ an existing UD parsing framework --- UDify, a multi-task learning model \citep{kondratyuk2019} --- and extend it to perform few-shot model adaptation to previously unseen languages via meta-learning.
We pre-train the dependency parser on a high-resource language prior to applying the model-agnostic meta-learning (\textsc{maml}) algorithm \citep{finn2017} to a collection of few-shot tasks in a diverse set of languages. We evaluate our model on its ability to perform few-shot adaptation to unseen languages, from as few as 20 examples. Our results demonstrate that our methods outperform language transfer and multilingual joint learning baselines, as well as existing (zero-shot) UD parsing approaches, on a range of language families, with the most notable improvements among the low-resource languages. We also investigate the role of a pre-training language as a starting point for cross-lingual adaptation and the effect of typological properties on the learning process.

\section{Related work}
\label{section:related_work}
\subsection{Meta-learning}

In meta-learning, the datasets are separated into \textit {episodes} that correspond to training tasks. Each episode contains a \textit{support} and a \textit{query} set, that include samples for adaptation and evaluation, respectively.
Meta-learning serves as an umbrella term for algorithms from three categories: 
\textit{Metric-based} methods classify new samples based on their similarity to the support set \citep[e.g.][]{snell2017prototypical}.
\textit{Model-based} methods explicitly store meta-knowledge within their architectures -- e.g. through an external memory \citep{santoro2016meta}.
\textit{Optimization-based} methods, on which we focus, estimate parameter initializations that can be fine-tuned with a few steps of gradient descent \citep[e.g.][]{finn2017, nichol2018reptile}.
\citet{finn2017} proposed \textsc{maml} to learn parameter initializations that generalize well to similar tasks. During the \textit{meta-training} phase, \textsc{maml} iteratively selects a batch of episodes, on which it fine-tunes the original parameters given the support set in an \textit{inner learning loop}, and tests it on the query set.
The gradients of the query set with respect to the original parameters are used to update those in the \textit{outer learning loop}, such that these weights become a better parameter initialization over iterations.
Afterwards, during \textit{meta-testing}, one selects a support set for the test task, adapts the model using that set and evaluates it on new samples from the test task.
\textsc{maml} has provided performance benefits for cross-lingual transfer for tasks such as machine translation \citep{gu2018meta}, named entity recognition \citep{wu2020enhanced}, hypernymy detection \citep{yu2020hypernymy} and mapping lemmas to inflected forms \citep{kann2020learning}. The closest approach to ours is by \citet{nooralahzadeh2020}, who focus on natural language inference and question answering. Their method, \textsc{x-maml}, involves pre-training a model on a high-resource language prior to applying \textsc{maml}. This yielded performance benefits over standard supervised learning for cross-lingual transfer in a zero-shot and fine-tuning setup
(albeit using 2500 training samples to fine-tune on test languages). The performance gains were the largest for languages sharing morphosyntactic features.
Besides the focus on dependency parsing, our approach can be distinguished from \citet{nooralahzadeh2020} in several ways. We focus on fast adaptation from a small number of examples (using only 20 to 80 sentences).
Whilst they use one language for meta-training, we use seven languages, with the aim of explicitly learning to adapt to a variety of languages.
\subsection{Universal dependency parsing}
The Universal Dependencies project is an ongoing community effort to construct a cross-linguistically consistent morphosyntactic annotation scheme \citep{universaldependencies}. 
The project makes results comparable across languages and eases the evaluation of cross-lingual (structure) learning.
The task of dependency parsing involves predicting a dependency tree for an input sentence, which is a directed graph of binary, asymmetrical arcs between words. These arcs are labeled and denote dependency relation types, which hold between a \textit{head}-word and its \textit{dependent}.
A parser is tasked to assign rankings to the space of all possible dependency graphs and to select the optimal candidate.

Dependency parsing of under-resourced languages has since long been of substantial interest in NLP.
Well-performing UD parsers, such as the winning model in the CoNLL 2018 Shared Task by \citet{che2018towards}, do not necessarily perform well on low-resource languages \citep{zeman2018conll}.
Cross-lingual UD parsing is typically accomplished by projecting annotations between languages with parallel corpora \citep{agic2014cross}, through model transfer \citep[e.g.][]{guo2015cross,ammar2016many,ahmad2018}, through hybrid methods combining annotation projections and model transfer \citep{tiedemann2014treebank}, or by aligning word embeddings across languages \citep{schuster2019cross}.

State-of-the-art methods for cross-lingual dependency parsing exploit pre-trained mBERT with a dependency parsing classification layer that is fine-tuned on treebanks of high-resource languages, and transferred to new languages:
\citet{wu2019} only fine-tune on English, whereas \citet{tran2019} experiment with multiple sets of fine-tuning languages. Including diverse language families and scripts benefits transfer to low-resource languages, in particular. UDify, the model of \citet{kondratyuk2019}, is jointly fine-tuned on data from 75 languages, with a multi-task learning objective that combines dependency parsing with predicting part-of-speech tags, morphological features, and lemmas. \citet{ustun2020udapter}, instead, freeze the mBERT parameters and train adapter modules that are interleaved with mBERT’s layers, and take a language embedding as input. This embedding is predicted from typological features. Model performance strongly relies on the availability of those features, since using proxy embeddings from different languages strongly degrades low-resource languages’ performance.

\section{Dataset}
\label{sec:dataset}
We use data from the Universal Dependencies v2.3 corpus \citep{universaldependencies}. We use treebanks from 26 languages that are selected for their typological diversity.
We adopt the categorization of high-resource and low-resource languages from \citet{tran2019} and employ their set of training and test languages for comparability. The set covers languages from six language families (Indo-European, Korean, Afro-Asiatic, Uralic, Dravidian, Austro-Asiatic).
Their training set ({\it expMix}) includes eight languages: English, Arabic, Czech, Hindi, Italian, Korean, Norwegian, and Russian. These languages fall into the language families of Indo-European, Korean and Afro-Asiatic and have diverse word orders (i.e. VSO, SVO and SOV).
Joint learning on data from this diverse set yielded state-of-the-art zero-shot transfer performance on low-resource languages in the experiments of \citet{tran2019}.

Per training language we use up to 20,000 example trees, predicting dependency arc labels from 132 classes total.
We select Bulgarian (Indo-European) and Telugu (Dravidian) as validation languages to improve generalization to multiple language families.
The 16 test languages cover three new language families that were unseen during training, i.e. Austro-Asiatic, Dravidian, and Uralic. Furthermore, three of our test languages (Buryat, Faroese, and Upper Sorbian) are not included in the pre-training of mBERT. We refer the reader to Appendix~\ref{ap:details:data} for details about the treebank sizes and language families.

\section{Method}
\label{section:method}
\subsection{The UDify model}
The UDify model concurrently predicts part-of-speech tags, morphological features, lemmas and dependency trees \cite{kondratyuk2019}. UDify exploits the pre-trained mBERT model \citep{devlin-etal-2019-bert}, that is a self-attention network with 12 transformer encoder layers.

The model takes single sentences as input. Each sentence is tokenized into subword units using mBERT's word piece tokenizer, after which contextual embedding lookup provides input for the self-attention layers. 
A weighted sum of the outputs of all layers is computed (\autoref{eq:UDify}) and fed to a task-specific classifier. 
\begin{equation}
\label{eq:UDify}
    e_{j}^{t} = \eta\sum_{i}\mathbf{B}_{ij}\cdot \text{softmax}{(\bm{\gamma})}_{i}
\end{equation}
Here, $e^t$ denotes the contextual output embeddings for task $t$. In our case, $t$ indicates UD-parsing. In contrast to the multi-task objective of the original UDify model, our experiments only involve UD-parsing.
The term $\mathbf{B}_{ij}$ represents the mBERT representation for layer $i=1,...,12$ at token position $j$.
The terms $\bm{\gamma}$ and $\eta$ denote trainable scalars, where the former applies to mBERT and the latter scales the normalized averages.
For words that were tokenized into multiple word pieces, only the first word piece was fed to the UD-parsing classifier. 

The UD-parsing classifier is a graph-based biaffine attention classifier \citep{dozat2016deep} that projects the embeddings $e_{j}^{t}$ through arc-head and arc-dep feedforward layers. The resulting outputs are combined using biaffine attention to produce a probability distribution of arc heads for each word. Finally, the dependency tree is decoded using the Chu-Liu/Edmonds algorithm \citep{chu1965shortest,edmonds1967optimum}. We refer the reader to the work of \citet{kondratyuk2019} for further details on the architecture and its training procedure.

\begin{figure}[t!]
  \centering
  \includegraphics[width=\columnwidth]{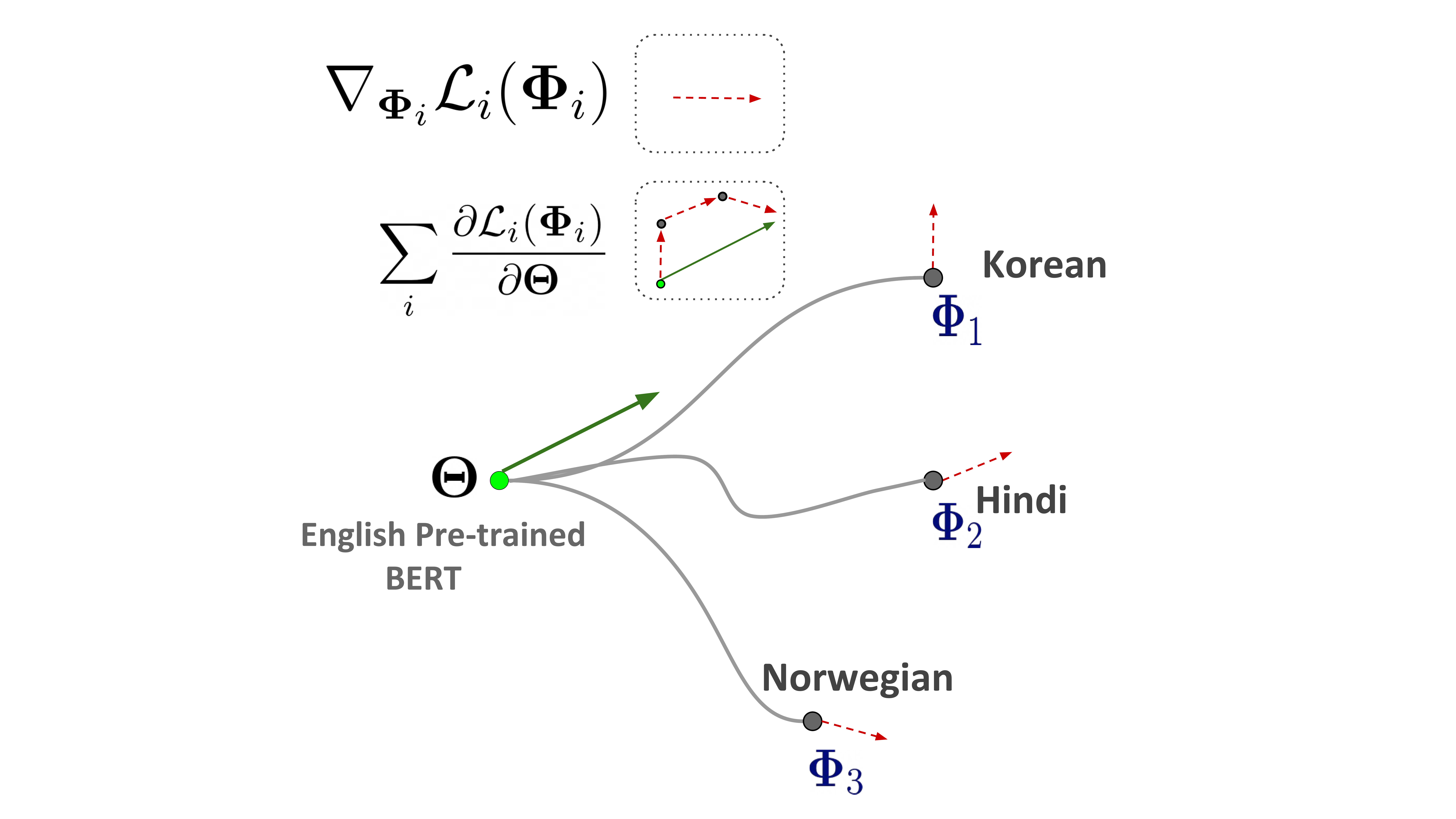}
  \caption{Visualization of \textsc{maml} algorithm for three meta-training languages. The green arrows represents the meta-update from the outer learning loop. The red dotted arrows represent the gradient computed on the support set for each language in the inner learning loop.}
  \label{fig:fomaml} 
\end{figure}

\subsection{Meta-learning procedure}
\label{sec:method:meta_learning_procedure}
We apply first-order\footnote{For more details on first-order versus second-order, see \citet{finn2017,shutova2020}.} \textsc{maml} to the UDify model. The model's self-attention layers are initialized with parameters from mBERT and the classifier's feedforward layers are randomly initialized. The model is pre-trained on a high-resource language using standard supervised learning and further meta-trained on a set of seven languages with \textsc{maml}. It is then evaluated using meta-testing. We refer to \textsc{maml} with pre-training as simply \textsc{maml}.
The meta-learning procedure is visualized in \autoref{fig:fomaml} and can be described as follows:
\paragraph{Step 1} Pre-train on a high-resource language to yield the initial parameters $\Theta$. 

\paragraph{Step 2} Meta-train on all other training languages. 
For each language $i$, we partition the UD training data into two disjoint sets, $D^{\text{train}}_{i}$ and $D^{\text{test}}_{i}$, and perform the following inner loop:
\begin{enumerate} 
    \item Temporarily update the model parameters $\Theta_i$ with stochastic gradient descent on support set $S$, sampled from $D^{\text{train}}_{i}$, with learning rate $\alpha$ for $k$ gradient descent adaptation steps. When using a single gradient step, the update becomes:
    \begin{equation}
        \Phi_i \leftarrow \Theta - \alpha \nabla_{\Theta}\mathcal{L}(\Theta_i)
        \end{equation}
    \item Compute the losses of the model parameters $\Phi_i$ using the query set $Q$, sampled from $D^{\text{test}}_{i}$, denoted by $\mathcal{L}_i(\Phi_i)$. 
\end{enumerate}
    
\paragraph{Step 3} Sum up the test losses and perform a meta-update in the outer learning loop on the model with parameters $\Theta$ using the learning rate $\beta$: 
\begin{equation}
    \Theta \leftarrow \Theta - \beta\sum_{i} \nabla_{\Theta} \mathcal{L}_i(\Phi_i)
\end{equation}
In our experiments, the update is a first-order approximation, replacing $\nabla_{\Theta} \mathcal{L}_i(\Phi_i)$ by $\nabla_{\Phi_i} \mathcal{L}_i(\Phi_i)$.
\paragraph{Step 4} After meta-training, we apply meta-testing to unseen languages. For each language, we sample a support set $S$ from the UD training data. We then fine-tune our model on $S$, and evaluate the model on the entire test set. 
Thereby, meta-testing mimics the adaptation from the inner loop. We repeat this process multiple times to get a reliable estimate of how well the model adapts to unseen languages.

\section{Experimental setup}
\label{section:setup}

We extend the existing UDify code\footnote{\href{https://www.github.com/Hyperparticle/udify}{github.com/Hyperparticle/udify}} to be used in a meta-learning setup. All of our code is publicly available.\footnote{\href{https://github.com/annaproxy/udify-metalearning}{github.com/annaproxy/udify-metalearning}}
\subsection{Training and evaluation }

\paragraph{Pre-training}
In the main body of the paper, we consider the pre-training languages English and Hindi to measure the impact of pre-training prior to cross-lingual adaptation, and to draw more general conclusions about how well \textsc{maml} generalizes with typologically different pre-training languages. 
English and Hindi differ in word order (SVO versus SOV), and
Hindi treebanks have a larger percentage of non-projective dependency trees \citep{mannem2009}, where dependency arcs are allowed to cross one another. Non-projective trees are more challenging to parse \citep{nivre2009}. Pre-training on Hindi allows us to test the effects of projectivity on cross-lingual adaptation.
To ensure that our findings are not specific to the pre-training languages of English and Hindi, \autoref{sec:czech} reproduces a subset of experiments for the pre-training languages Italian and Czech, reporting results for monolingual baselines, a non-episodic baseline, and \textsc{maml}. 
Italian and Czech are high-resource languages as well, but are from two different subfamilies of the family of Indo-European languages and also differ in their percentage of non-projective dependency trees.
\paragraph{Meta-training} 
We apply meta-training using seven languages listed in Section~\ref{sec:dataset},
excluding the pre-training language from meta-training. We train for 500 episodes per language, using a cosine-based learning rate scheduler with 10\% warm-up. We use the Adam optimizer \citep{kingma2014adam} in the outer loop and SGD in the inner loop \citep{finn2017}. Support and query sets are of size 20. Due to the sequence labelling paradigm, the number of shots per class varies per batch. When $|S|=20$, the average class will appear 16 times.
To select hyperparameters, we independently vary the amount of updates $k$ and the learning rates in the inner loop and outer loop for mBERT and the parser, while performing meta-validation with the languages Bulgarian and Telugu. To meta-validate, we follow the procedure described in Section~\ref{sec:method:meta_learning_procedure} for both languages, mimicking the meta-testing setup with a support set size of 20.
The hyperparameters are estimated independently for Hindi and English pre-training (see Appendix~\ref{sec:appendix_hyperparameters}).
\paragraph{Meta-testing} 
At meta-testing time, we use SGD with the same learning rates and the same $k$ used in the inner loop during meta-training. We vary the support set size $|S| \in \{20,40,80\}$.
\subsection{Baselines}
\label{sec:baselines}

We define several baselines, that are evaluated using meta-testing, i.e. by fine-tuning the models on a support set of a test language prior to evaluation on that language. This allows us to directly compare their ability to {\it adapt quickly} to new languages with that of the meta-learner. 

\paragraph{Monolingual baselines (\textsc{en}, \textsc{hin})}
These baselines measure the impact of meta-training on data from seven additional languages. The model is initialized using mBERT and trained using data from English (\textsc{en}) or Hindi (\textsc{hin}), without meta-training. 
\paragraph{Multilingual non-episodic baseline (\textsc{ne})}
Instead of episodic training, this baseline treats support and query sets as regular mini-batches and updates the model parameters directly using a joint learning objective, similar to \citet{kondratyuk2019} and \citet{tran2019}.
The model is pre-trained on English or Hindi and thus indicates the advantages of \textsc{maml} over standard supervised learning.
The training learning rate and meta-testing learning rate are estimated separately, since there is no inner loop update in this setup.

\paragraph{\textsc{maml} without pre-training} We evaluate the effects of pre-training by running a \textsc{maml} setup without any pre-training.
Instead, the pre-training language is included during meta-training as one of now eight languages. \textsc{maml} without pre-training is trained on 2000 episodes per language. 

\paragraph{Meta-testing only} The simplest baseline is a decoder randomly initialized on top of mBERT, without pre-training and meta-training. Dependency parsing is only introduced at meta-testing time.

\subsection{Evaluation}
Hyperparameter selection and evaluation is performed using Labeled Attachment Score (LAS) as computed by the CoNLL 2018 Shared Task evaluation script.\footnote{\href{https://universaldependencies.org/conll18/evaluation.html}{universaldependencies.org/conll18/evaluation.html}} 
LAS evaluates the correctness of both the dependency class and dependency head. We use the standard splits of Universal Dependencies for training and evaluation when available. Otherwise, we remove the meta-testing support set from the test set prior to evaluation. We train each model with seven different seeds and compare \textsc{maml} to a monolingual baseline and \textsc{ne} using paired $t$-tests, adjusting for multiple comparisons using Bonferroni correction.

\section{Results and analysis}
\label{section:results}
\begin{table*}[t!]
  \centering\small
  \begin{tabular}{lccc|ccc|ccc|ccc}\toprule
& & {\bf } && \multicolumn{3}{c|}{$|S|=20$} & \multicolumn{3}{c|}{$|S|=40$} & \multicolumn{3}{c}{$|S|=80$} \\
\textbf{Language} & \textbf{T\&B} & \textbf{K\&S} & \textbf{\"{U}st.} & \textbf{\textsc{en}} & \textbf{\textsc{ne}} & \textbf{\textsc{maml}} & \textbf{\textsc{en}} & \textbf{\textsc{ne}} & \textbf{\textsc{maml}} & \textbf{\textsc{en}} & \textbf{\textsc{ne}} & \textbf{\textsc{maml}} \\ \midrule
\multicolumn{4}{l|}{\textit{Low-Resource Languages}} & \multicolumn{3}{l|}{} & \multicolumn{3}{l|}{} \\
Armenian & 58.95  &  --  & --  &  49.8  & 63.34 &\underline{\bf 63.84} & 50.59 & 63.54 & \underline{\bf 64.30} & 51.99 & 63.79 & \underline{\bf 64.78} \\
Breton & 52.62    & 39.84 & 58.5 &   60.34  & 61.44 &\underline{\bf 64.18} & 61.32 & 61.67 & \underline{\bf 65.12} & 62.76 & 62.20 & \underline{\bf 66.14} \\
Buryat$\dagger$   & 23.11   &  26.28 & \bf 28.9  & 23.66 & 25.56 & {\bf 25.77} & 23.82 & 25.67 & \underline{\bf 26.38} & 24.17 & 25.88 & \underline{\bf 27.33} \\
Faroese$\dagger$  & 61.98   &  59.26 &  69.2  & 68.50 & 67.83 & {\bf 68.95} & 69.56 & 68.12 & \underline{\bf 69.88} & 70.59 & 68.62 & \underline{\bf 71.12} \\
Kazakh & 44.56    &\bf 63.66& \bf 60.7 &  47.25  & 55.02 &\bf 55.07 & 47.80 & 55.08 & \underline{\bf 55.46} & {49.08} & 55.23 & \underline{\bf 56.15} \\
U.Sorbian$\dagger$& 49.74   &{\bf 62.82} & 54.2& 49.29 & 54.47 & \underline{\bf 56.40} & {50.55} & 54.70 & \underline{\bf 57.55} & {52.11} & {55.07} & \underline{\bf 58.81} \\
\textit{Mean}     & 48.45   & -- & -- & 49.81 & 54.61 & 55.70 & 50.61 & 54.80 & 56.45 & 51.78 & 55.13 & 57.38\\ 
\midrule
\multicolumn{4}{l|}{\textit{High-Resource Languages}} & \multicolumn{3}{l|}{} & \multicolumn{3}{l|}{} \\
Finnish           & 62.29   & --  & --& 56.61 & \bf 64.94 & {64.89} & {56.99} & {65.07} & {\bf 65.40} & {57.73} & {65.18} & \underline{\bf 65.82} \\
French 	          & 59.54   & --  & --& 65.21 & 66.55 & \underline{\bf 66.85} & {65.33} & {66.59} & \underline{\bf 66.97} & {65.63} & {66.65} & \underline{\bf 67.25} \\
German            & 70.93   & -- & -- & 72.47 & 76.15 & {\bf 76.41} & {72.6} & {76.17} & {\bf 76.54} & {72.93} & {76.21} & \underline{\bf 76.72} \\
Hungar.         & 61.11   & -- & -- & 56.50 &\bf 62.93 & {62.71} & {56.23} & {\bf 63.09} & {62.81} & {56.73} & \underline{\bf 63.21} & 62.52 \\
Japanese          & 24.10   & -- & -- & 18.87 & 36.49 & \underline{\bf 39.06} & {20.05} & {37.15} & \underline{\bf 42.17} & 22.80 & 38.40 & \underline{\bf 46.81} \\
Persian           &\bf 56.92& --  & --& 43.43 & 52.55 & {\bf 52.81} & {44.53} & {52.76} & {\bf 53.63} & {46.42} & {53.11} & \underline{\bf 54.74} \\
Swedish           & 78.70   & --  & --& 80.26 & 80.73 & {\bf 81.36} & {80.41} & {80.81} & \underline{\bf 81.53} & {80.57} & {80.79} & \underline{\bf 81.59} \\
Tamil             & 32.78   & --  & -- & 31.58 & 41.12 & \underline{\bf 44.34} & {32.67} & {41.72} & \underline{\bf 46.73} & {34.81} & {42.88} & \underline{\bf 50.73} \\
Urdu              &\bf 63.06& -- & -- & 25.71 & \underline{\bf 57.25} & {55.16} & {26.89} & \underline{\bf 57.36} & {56.16} & 29.30 & {\bf 57.68} & 57.60 \\
Vietnam.          & 29.71   & --  & -- & 43.24 & {42.73} & {\bf 43.34} & {43.65} & {42.82} & {\bf 43.74} & {44.28} & {43.02} & {\bf 44.34} \\
\textit{Mean} &53.91 & --  & --&  49.39 & 58.14 & 58.69 & 49.93 & 58.35 & 59.57 & 51.12 & 58.71 & 60.81 \\
\midrule 
\textit{Mean}  & 51.88 &  -- & -- & 49.55 & 56.82 & 57.57 & 50.19 & 57.02 & 58.4 & 51.37 & 57.37 & 59.52\\
\bottomrule
\end{tabular}

  \caption{
  Mean LAS aligned accuracy per support set size $|S|$ for unseen test languages. Best results per category are bolded. Significant results are underlined ($p<0.005$). 
  Previous work consists of \citet{tran2019}, UDify~\citep{kondratyuk2019} and UDapter~\citep{ustun2020udapter}.
  $\dagger$: Languages were absent from mBERT.
  }
  \label{tab:testresults}
\end{table*}

\paragraph{\textsc{maml} with English pre-training} We report the mean LAS for models pre-trained on English in \autoref{tab:testresults}.
We compare these results to related approaches that use mBERT and have multiple training languages.
With support set size 20, \textsc{maml} already outperforms the zero-shot transfer setup of \citet{tran2019} for all test languages, except Persian and Urdu.
\textsc{maml} is competitive with UDify \citep{kondratyuk2019} and UDapter \citep{ustun2020udapter} for low-resource languages, despite the stark difference in the number of training languages compared to UDify\footnote{UDify is trained on the low-resource languages, while we only test on them. For a fair comparison, we only list UDify results on languages with a small amount of sentences ($<$80) in the training set, to mimic a few-shot generalisation setup.} (75), and without relying on fine-grained typological features of languages, as is the case for UDapter.

\textsc{maml} consistently outperforms the \textsc{en} and \textsc{ne} baselines. Large improvements over the \textsc{en} baseline are seen on low-resource and non-Germanic languages.
The difference between \textsc{maml} and the baselines increases with $|S|$. The largest improvements over \textsc{ne} are on Tamil and Japanese, however \textsc{ne} outperforms \textsc{maml} on Hungarian and Urdu. \textsc{maml} consistently outperforms \textsc{ne} on low-resource languages, with an average 1.1\% improvement per low-resource language for $|S|=20$, up to a 2.2\% average improvement for $|S|=80$.

\begin{table}[t!]
  \centering\small
  \begin{tabular}{l|rr|rr}
\toprule
& \multicolumn{2}{c|}{$|S| = 20$} & \multicolumn{2}{c}{$|S| = 80$} \\
\textbf{Language} & \textbf{\textsc{maml}} & \textbf{\MAMLPrime} & \textbf{\textsc{maml}} & \textbf{\MAMLPrime} \\
\midrule
\multicolumn{3}{l|}{\textit{Low-Resource Languages}} & \multicolumn{2}{l}{}  \\
Armenian        &\bf\underline{63.84} & 59.70 & \bf\underline{64.78} & 60.03 \\
Breton          &\bf\underline{64.18} & 59.33 & \bf\underline{66.14} & 60.84 \\
Buryat$\dagger$ & 25.77 & \bf 26.02 & \bf 27.33 & 27.05 \\
Faroese$\dagger$&\bf\underline{68.95}& 65.30  & \bf\underline{71.12} & 66.79 \\
Kazakh          &\bf\underline{55.07} & 53.92 & \bf\underline{56.15} & 54.99 \\
U.Sorbian$\dagger$&\bf\underline{56.40} & 51.67  & \bf\underline{58.78} & 52.38 \\
\it Mean & 55.7 & 52.66 & 57.38 & 53.68 \\
\midrule
\multicolumn{3}{l|}{\textit{High-Resource Languages}} & \multicolumn{2}{l}{} \\
\it Mean & 58.69 & 57.04  & 60.81 & 58.25 \\ 
\bottomrule
\end{tabular}

  \caption{Mean LAS per unseen language, for \textsc{maml} without pre-training (denoted \MAMLPrime{}) versus \textsc{maml} (EN). $\dagger$: Languages were absent from mBERT.}
  \label{tab:mamlresults}
\end{table}

\begin{table*}[t!]
  \centering\small
  \begin{tabular}{l|rrr|rrr|rrr}
\toprule
 & \multicolumn{3}{c|}{$|S|=20$} & \multicolumn{3}{c|}{$|S|=40$} & \multicolumn{3}{c}{$|S|=80$} \\
\textbf{Language}  & \textbf{\textsc{hin}} & \textbf{\textsc{ne}} & \textbf{\textsc{maml}} & \textbf{\textsc{hin}} & \textbf{\textsc{ne}} & \textbf{\textsc{maml}} & \textbf{\textsc{hin}} & \textbf{\textsc{ne}} & \textbf{\textsc{maml}} \\ \midrule
\multicolumn{4}{l|}{\textit{Low-Resource Languages}} & \multicolumn{3}{l|}{}  \\
Armenian  		  & 48.41 & 63.30 & \bf 63.76  & 48.87 & 63.41 & \bf 64.17 & 49.70 & 63.59 & \bf\underline{64.76} \\
Breton  		  & 34.06 & \bf 62.09 & 61.56  & 36.09 & 62.40 & \bf 62.47 & 38.95 & 63.05 & \bf 63.75  \\
Buryat$\dagger$   & 24.24 & 25.05 & \bf\underline{26.27} & 24.71 & 25.18 & \bf 26.79 & 25.54 & 25.40 & \bf\underline{27.37} \\
Faroese$\dagger$  & 50.72 & 65.31 & \bf 66.82  & 52.30 & 65.57 & \bf\underline{67.31}& 54.64 & 66.17 & \bf\underline{68.25} \\
Kazakh  		  & 49.80 & 53.77 & \bf\underline{54.23} & 49.90 & 53.94 & \bf\underline{54.45}& 50.49 & 54.08 & \bf\underline{55.00} \\
U.Sorbian$\dagger$& 36.22 & 53.36 & \bf 54.97  & 37.08 & 53.58 & \bf\underline{55.64}& 38.22 & 53.94 & \bf 56.56  \\
\textit{Mean}     & 40.57 & 53.81 & 54.60      & 41.49 & 54.01 & 55.14     & 42.92 & 54.37 & 55.95      \\ \midrule
\multicolumn{4}{l|}{\textit{High-Resource Languages}} & \multicolumn{3}{l|}{}  \\
Finnish           & 50.49 & 64.05 & \bf 64.64 & 50.93 & 64.20 & \bf\underline{65.05} & 51.79 & 64.40 & \bf\underline{65.61} \\
French            & 31.16 & 64.44 & \bf\underline{65.73}& 31.59 & 64.44 & \bf\underline{65.68} & 33.39 & 64.42 & \bf\underline{65.69} \\
German            & 44.83 & 74.40 & \bf 75.15 & 45.46 & 74.41 & \bf\underline{75.23} & 46.65 & 74.46 & \bf\underline{75.31} \\
Hungarian         & 46.72 & 60.98 & \bf 62.51 & 46.97 & 61.33 & \bf\underline{62.89} & 47.91 & 61.68 & \bf\underline{62.91} \\
Japanese          & 40.25 & 39.97 & \bf\underline{41.96}& 43.03 & 40.56 & \bf 43.61  &\bf 46.87 & 41.58 & 45.90 \\
Persian           & 28.60 & \bf 53.73 & 53.63 & 29.51 & 53.85 & \bf 54.00  & 31.11 & 54.06 & \bf 54.53 \\
Swedish           & 46.96 & 79.24 & \bf\underline{79.89}& 47.73 & 79.32 & \bf\underline{80.14} & 49.15 & 79.31 & \bf\underline{80.21} \\
Tamil             &\bf\underline{46.51}& 39.44 & 39.57&\bf\underline{47.35}& 39.84& 40.84 & \bf\underline{48.55} & 40.73 & 42.81 \\
Urdu              &\bf\underline{67.72}& 50.64 & 49.16&\bf\underline{67.96}& 50.93& 50.16 & \bf\underline{68.17} & 51.50 & 51.57 \\
Vietnamese        & 26.96 & \bf 42.13& 42.12& 27.92   & 42.23 & \bf 42.37 & 29.61 & 42.46 & \bf 42.87 \\
\textit{Mean}     & 43.02 & 56.9 & 57.44 & 43.85 & 57.11 & 58.0 & 45.32 & 57.46 & 58.74\\
\midrule 
\textit{Mean} & 42.1 & 55.74 & 56.37 & 42.96 & 55.95 & 56.92 & 44.42 & 56.3 & 57.69\\
\bottomrule 
\end{tabular}

  \caption{Mean LAS aligned accuracy per unseen language, for models pre-trained on Hindi. Best results per category are listed in bold, significant results are underlined ($p<0.005$). $\dagger$: Languages were absent from mBERT.}
  \label{tab:hindi_testresults}
\end{table*}

\paragraph{\textsc{maml} with Hindi pre-training} The results for models pre-trained on Hindi can be seen in \autoref{tab:hindi_testresults}. Although there are large differences between the monolingual \textsc{en} and \textsc{hin} baselines, both \textsc{maml} (\textsc{hin}) and \textsc{ne} (\textsc{hin}) achieve, on average, similar LAS scores to their English counterparts.
\textsc{maml} still outperforms \textsc{ne} for the majority of languages: the mean improvement on low-resource languages is 0.8\% per language for $|S|=20$, which increases to 1.6\% per language for $|S|=80$. 

\paragraph{Other pre-training languages} The full results for the two other pre-training languages, Italian and Czech, are listed in \autoref{sec:czech}. Here, too, \textsc{maml} outperforms its \textsc{ne} counterpart. The \textsc{ne} baseline is stronger for more languages than in our main experiments. For $|S|=20$, the mean improvements per unseen language are 0.91\% and 0.47\% when pre-training on Italian and Czech, respectively. For $|S|=80$, the improvements are 2.18\% and 1.75\%. 

\paragraph{\textsc{maml} without (pre-)training} We investigate the effectiveness of pre-training by omitting the pre-training phase. 
A comparison between \textsc{maml} and \textsc{maml} without pre-training is shown in \autoref{tab:mamlresults}.
\textsc{maml} without pre-training underperforms for most languages and its performance does not increase as much with a larger support set size. This suggests that pre-training provides a better starting point for meta-learning than plain mBERT.

When meta-testing only -- i.e. omitting both pre-training and meta-training -- the fine-tuned model reaches a mean LAS of 6.9\% over all test languages for $|S|=20$, increasing to 15\% for $|S|=80$, indicating that meta-testing alone is not sufficient to learn the task.\footnote{Full results can be found in \autoref{ap:fullresults}.}

\paragraph{Further Analysis}
Performance increases over the monolingual baselines vary strongly per language -- e.g. consider the difference between Japanese and French in \autoref{tab:testresults}. 
The performance increase is largest for languages that differ from the pre-training language with respect to their syntactic properties. 
 We conduct two types of analysis, based on typological features and projectivity,
to quantify this effect and correlate these properties to the performance increase over  monolingual baselines.\footnote{No clear correlation was found by \citet{tran2019}.
By using {\it increase in performance} and not ``plain" performance, we may see a stronger effect.}

\begin{figure}[!tb]
    \centering
    \includegraphics[width=0.75\columnwidth]{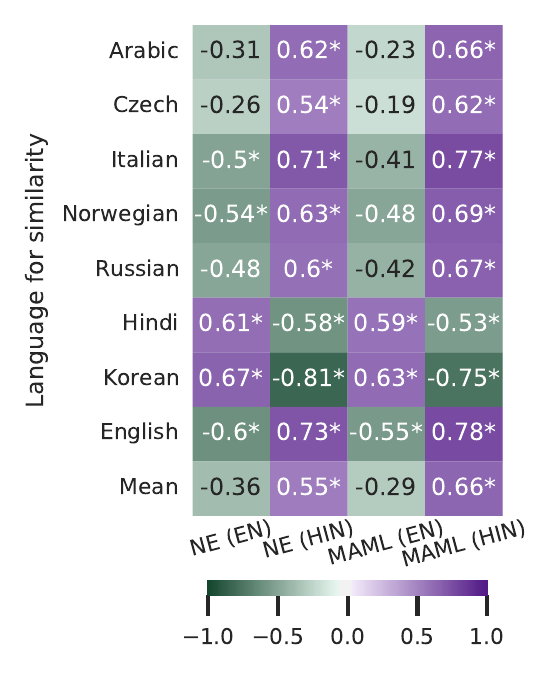}
    \caption{Spearman's $\rho$ between the performance increase over the monolingual baseline and the cosine similarity to the syntax of training languages ($y$-axis) for models using pre-training ($x$-axis). (*: $p <0.05$)}
    \label{fig:syntaxcorrelation}
\end{figure}

Firstly, we use 103 binary syntactic features from URIEL \citep{littel2017} to compute the syntactic cosine similarities (denoted $\sigma$) between languages. 
With this metric, a language such as Italian is syntactically closer to English ($\sigma=0.86$) than Urdu ($\sigma=0.62$), even though they are both Indo-European.
For each unseen language, we collect the cosine similarities to each (pre-)training language.
Then, we collect the difference in performance between the monolingual baselines and the \textsc{ne} or \textsc{maml} setups for $|S|=20$. 
For each training language, we compute the correlations between performance increases for the test languages and their similarity to this training language, visualised in \autoref{fig:syntaxcorrelation}.
When pre-training on Hindi, there is a significant positive correlation with syntactic similarity to English and related languages. 
When pre-training on English, a positive correlation is seen with similarity to Hindi and Korean. Positive correlations imply that on unseen languages, improvement increases when similarity to the training language increases. Negative correlations mean there is less improvement when similarity to the training languages increases, suggesting that those languages do not contribute as much to adaptation. 
On average, the selection of meta-training languages contributes significantly to the increase in performance for the Hindi pre-training models. This effect is stronger for \textsc{maml} (HIN) ($p=0.006$) than \textsc{ne} (HIN) ($p=0.026$), which may indicate that the meta-training procedure is better at incorporating knowledge from those unrelated languages. 

\begin{figure}[!t]
    \centering
    \includegraphics[width=0.9\columnwidth]{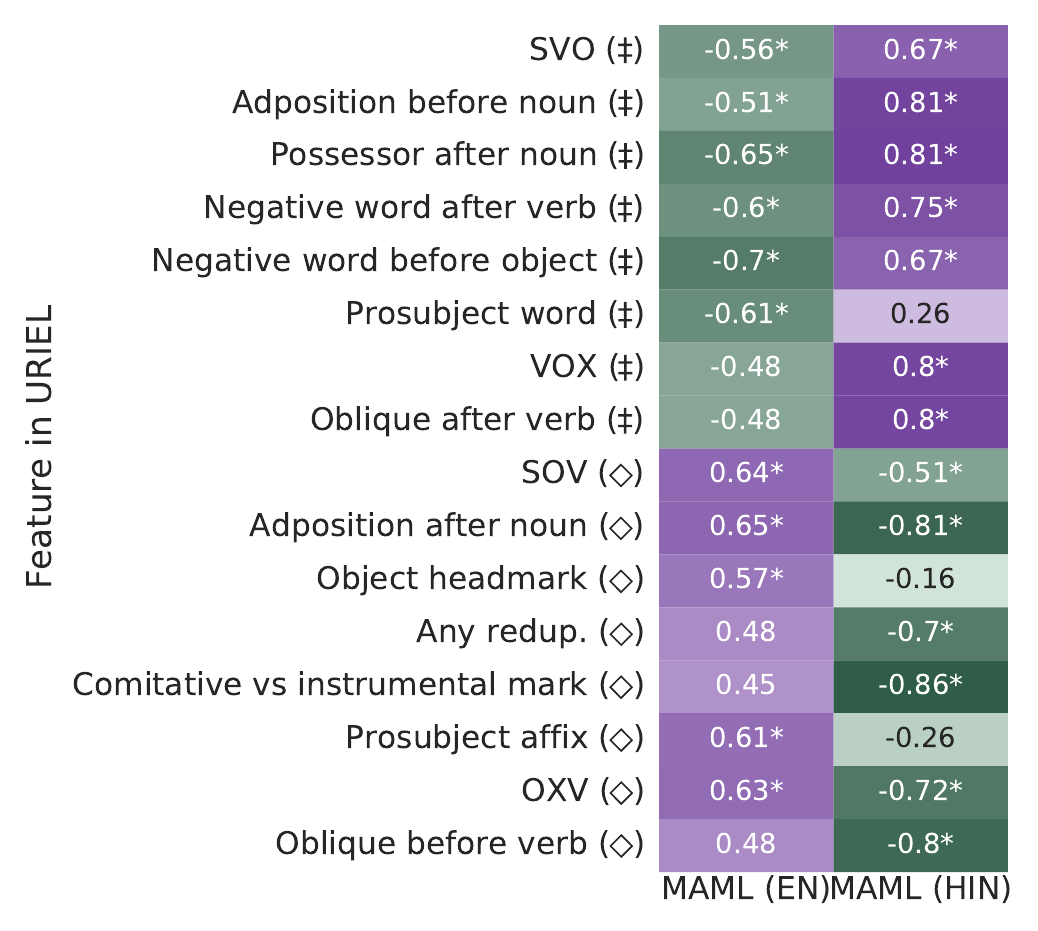}
    \caption{Spearman's $\rho$ between the performance increase over monolingual baselines and URIEL features ($y$-axis), for \textsc{maml} ($x$-axis). We indicate features present in English ($\ddagger$) and in Hindi ($\diamond$). (*: $p <0.05$)}
    \label{fig:urielcorrelation}
\end{figure}

Secondly, we analyze which syntactic features impact performance most. We correlate individual URIEL features with \textsc{maml}'s performance increases over monolingual baselines (see \autoref{fig:urielcorrelation}).
Features related to word order and negation show a significant correlation.
Considering the presence of these features in both pre-training languages of \textsc{maml}, a pattern emerges: when a feature is absent in the pre-training language, there is a positive correlation with increase in performance. Similarly, when a feature is present in the pre-training language, there is a negative correlation, and thus a smaller increase in performance after meta-training.
This indicates that \textsc{maml} is successfully adapting to these specific features during meta-training.

\begin{figure}[!t]
    \centering
    \includegraphics[width=0.8\columnwidth]{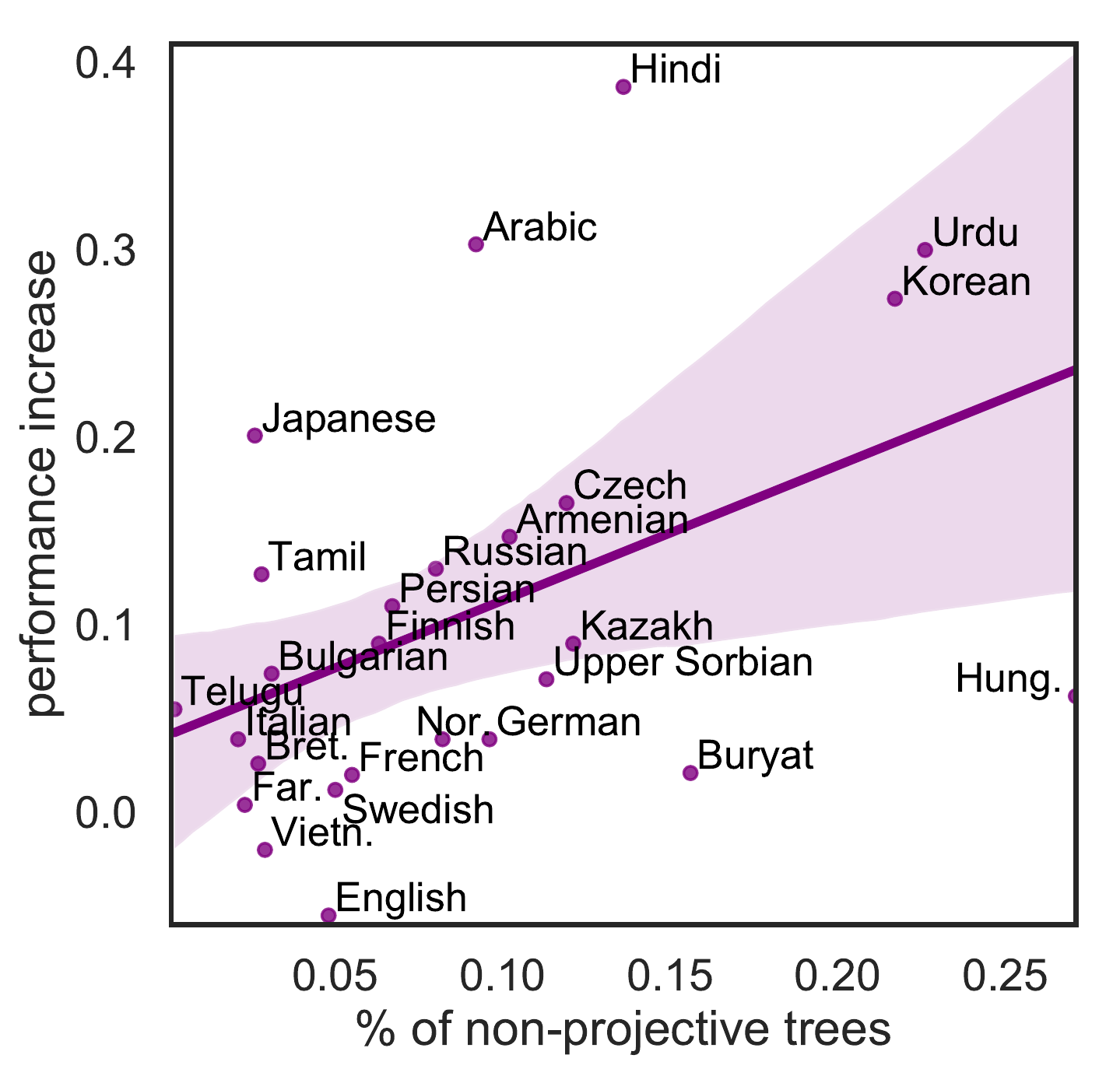}
    \caption{Spearman's $\rho$ between the \% of non-projective dependency trees and \textsc{maml}'s improvement over the English baseline ($\rho$ = $0.42$, $p$ = $0.03$).}
    \label{fig:projectivity}
\end{figure}

We analyzed \textsc{maml}'s performance improvements over \textsc{ne} on each of the 132 dependency relations, and found that they are consistent across relations.
 \footnote{The same holds for the 37 coarse-grained UD relations.} 
Lastly, we detect non-projective dependency trees in all datasets. The Hindi treebank used has 14\% 
of non-projective trees, whereas English only has 5\%.\footnote{Full results can be found in \autoref{ap:details:data}.} We correlate the increase in performance with the percentage of non-projective trees in a language's treebank. The correlation is significant for \textsc{ne} (EN) ($\rho$ = $0.46$, $p$ = $0.01$) and \textsc{maml} (EN) ($\rho$ = $0.42$, $p$ = $0.03$). \autoref{fig:projectivity} visualizes the correlation for \textsc{maml} (EN). We do not find significant correlations for models pre-trained on Hindi. This suggests that a model trained on a mostly projective language can benefit more from further training on non-projective languages than the other way around. The same trend is observed when comparing models pre-trained on Italian and Czech, that also differ in the percentage of non-projective trees (\autoref{sec:czech}).

\section{Discussion}
\label{section:discussion}
Our experiments confirm that meta-learning, specifically \textsc{maml}, is able to adapt to unseen languages on the task of cross-lingual dependency parsing more effectively than a non-episodic model. 
The difference between both methods is most apparent for languages that differ strongly from those in the training set (e.g. Japanese in \autoref{tab:testresults}) 
 where effective few-shot adaptation is crucial. 
This shows that \textsc{maml} is successful at \emph{learning to learn} from a few examples, and can efficiently incorporate new information. Furthermore, we see a clear increase in performance for \textsc{maml} when increasing the test support set size, while \textsc{ne} only slightly improves. This suggests that \textsc{maml} may be a promising method for cross-lingual adaptation more generally, also outside of the few-shot learning scenario.

Our ablation experiments on pre-training show that it is beneficial for \textsc{maml} to start from a strong set of parameters, pre-trained on a high resource language. 
Thereby, the pre-training is not dependent on a specific language. \textsc{maml} performs well with a variety of pre-training languages, although improvements for unseen languages vary. 
When a model is pre-trained on English, there is a large positive correlation for improvements in languages that are syntactically dissimilar to English, such as Japanese and Tamil. 
During meta-training, dissimilar training languages such as Hindi most contribute to the model's ability to generalize. 
Syntactic features, especially those related to word order, which have already been learned during pre-training, require less adaptation.
The same is true, vice versa, for Hindi pre-training.

This effect is also observed, though only in one direction, when correlating performance increase with non-projectivity. 
It is beneficial to meta-train on a set of languages that vary in projectivity after pre-training on one which is mostly projective.
However, not all variance is explained by the difference in typological features. 
The fact that \textsc{maml} outperforms \textsc{maml} without pre-training suggests that pre-training also contributes \emph{ language-agnostic} syntactic features, which is indeed the overall goal of multi-lingual UD models.

\section{Conclusion}
\label{section:conclusion}
In this paper, we present a meta-learning approach for the task of cross-lingual dependency parsing. 
Our experiments show that meta-learning can improve few-shot universal dependency parsing performance on unseen, unrelated test languages, including low-resource languages and those not covered by mBERT.
In addition, we see that it is beneficial to pre-train before meta-training, as in the \textsc{x-maml} approach \citep{nooralahzadeh2020}. In particular, the pre-training language can affect how much adaptation is necessary on languages that are typologically different from it. 
Therefore, an important direction for future research is to investigate a wider range of pre-training/meta-training language combinations,
based on specific hypotheses about language relationships and relevant syntactic features. 
 Task performance may be further improved by including a larger set of syntax-related tasks, such as POS-tagging, to sample from during meta-training \citep{kondratyuk2019}.

\bibliography{bibliography_rebibed}

\clearpage \newpage
\appendix 
\section{Training details and hyperparameters}
\label{sec:appendix_hyperparameters}

\begin{table}[h]
  \centering\small
  \begin{tabular}{c|c}
    \toprule 
    Parameter & Value \\
    \midrule 
    Dependency tag dimension & 256 \\
    Dependency arc dimension &  768 \\ 
    Dropout & 0.5 \\ 
    BERT Dropout & 0.2 \\ 
    Mask probability & 0.2 \\ 
    Layer dropout & 0.1 \\ 
    \bottomrule
  \end{tabular}
  \caption{Hyperparameters for the UDify model architecture.}
  \label{tab:udify_hyper}
\end{table}
All models use the same architecture: an overview of all model parameters can be seen in \autoref{tab:udify_hyper}. The model contains 196M parameters, of which 178M are mBERT. 

At pre-training time, we use the default parameters of UDify \citep{kondratyuk2019}. We pre-train for 60 epochs. The Adam optimizer is used with a 1e-3 learning rate for the decoder and a 5e-5 learning rate for BERT layers. Weight decay of 0.01 is applied. We employ a gradual unfreezing scheme, freezing the BERT layer weights for the first epoch. 

At meta-training time, we vary the learning rates as shown in \autoref{tab:lrs}. We also vary the amount of updates $k$ at training/testing time: $k \in \{8,20\}$. We applied weight decay at meta-training time in initial experiments, but this yielded no improvements. No gradual unfreezing is applied at meta-training time. We use Adam for the outer loop updates and SGD for the inner loop updates and at testing time. We sample 500 episodes per language, using query and support set size of 20. The best hyperparameters are chosen with respect to their final performance on the meta-validation set consisting of Bulgarian and Telugu. We run two seeds for hyperparameter selection, and seven seeds for all the final models. Labeled Attachment Score (LAS) is used for hyperparameter selection and final evaluation.

We train all models on an NVIDIA TITAN RTX. Pre-training takes around 3 hours, meta-training takes around 1 hour for 100 episodes per language when the amount of updates $k$ is set to 20. For \textsc{maml}, this amounts to approximately 5 hours, and for the ablated \MAMLPrime{}, it amounts to approximately 20 hours, which can be seen as another benefit of pre-training. Finally, training in a non-episodic fashion (\textsc{NE}) also takes up less time, namely 2 to 3 hours. 

All final learning rates can be seen in \autoref{tab:finals}. For all models except the random decoder baseline, $k=20$ was selected. The best random decoder used $k=80$  after a separate hyperparameter search of high learning rates and $k$s (compensating for the lack of prior DP training).
\begin{table}[ht]
  \centering\small
  \begin{tabularx}{\linewidth}{lrX}\toprule
    \bf LR & \bf mBERT & \bf Decoder \\ \midrule
    Inner/Test $\alpha$ & \{1e-4, \underline{5e-5}, \underline{1e-5}\} & \{\underline{1e-3}, 5e-4, \underline{1e-4}, 5e-5\} \\
    Outer $\beta$ & \{{\underline{5e-5}}, \underline{1e-5}, 7e-6\} & \{\underline{1e-3}, \underline{7e-4}, \underline{5e-4}, \underline{1e-4}, 5e-5\}\\ 
    \bottomrule
  \end{tabularx}
  \caption{Learning rates independently varied for \textsc{maml} and \textsc{ne}. For the ablated \MAMLPrime{}, only \underline{underlined} learning rates were tried due to the long training times.}
  \label{tab:lrs}
\end{table}

\begin{table}[ht]
  \centering\small
  \begin{tabular}{l|rr|rr}
  \toprule
    & \multicolumn{2}{c}{\bf Inner/Test LR} & \multicolumn{2}{c}{\bf Outer LR}\\
    & Decoder & BERT & Decoder & BERT \\
    \midrule
    Meta-test only & 5e-3 & 1e-3  & n/a & n/a  \\
    \textsc{en}/\textsc{hin} & 1e-4  &1e-4  & n/a & n/a  \\
    \textsc{ne} (en/hin) & 5e-4 & 1e-5 & 1e-4 & 7e-6 \\
    \textsc{maml} (en) & 1e-3 & 1e-4 & 5e-4 & 1e-5 \\
    \textsc{maml} (hin) & 5e-4 & 5e-5 & 5e-4 &5e-5 \\
    \MAMLPrime{} & 1e-3 & 1e-5 & 5e-4 & 1e-5 \\
    \bottomrule
  \end{tabular}
  \caption{Final hyperparameters, as selected by few-shot performance on the validation set. Inner loop/Test learning rates are used with SGD, outer loop LRs are used with the Adam optimizer.}
  \label{tab:finals}
\end{table}

\section{Information about datasets used}
\label{ap:details:data}
\begin{figure*}[!h]
  \centering
  \includegraphics[width=\textwidth]{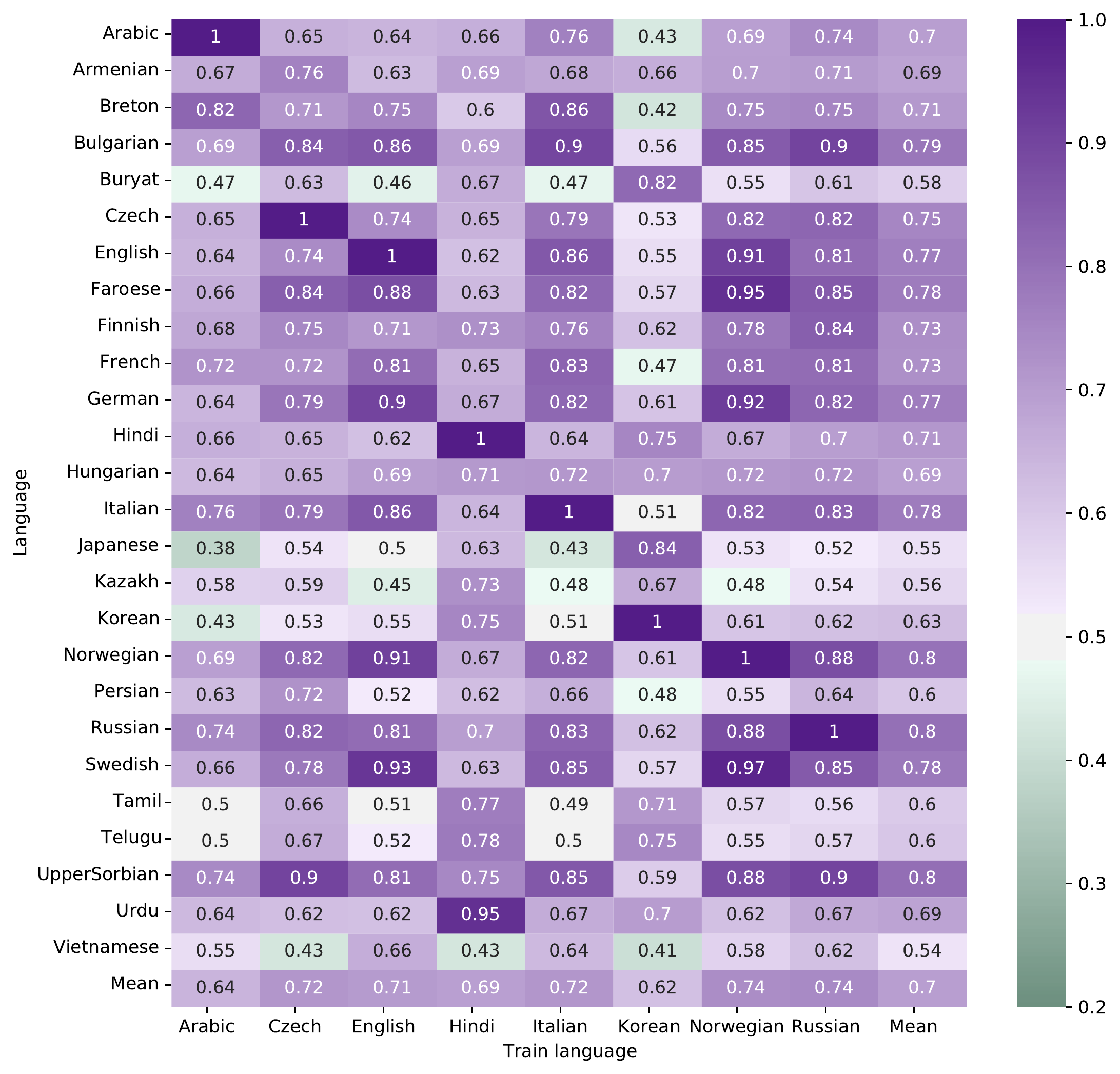}
  \caption{Syntactical cosine similarities from each training language to all other languages, calculated using URIEL's 103 binary syntactic features \citep{littel2017}. Average cosine similarities are shown in the rightmost column and the bottom row. For instance, Japanese and Kazakh have a relatively low average cosine similarity to the training languages. }
  \label{fig:gridsimilarities}
\end{figure*}

All information about the datasets used can be found in \autoref{tab:all_dataset_info}, along with corresponding statistics about non-projective trees. The cosine syntactical similarities are visualized in \autoref{fig:gridsimilarities}.

\section{Full results}
\label{ap:fullresults}
\begin{table*}[!t]
  \centering\small
 \begin{tabular}{llllrrrr}
    \toprule
    \bf Language & \bf Family & \bf Subcategory & \bf UD Dataset & \bf Train & \bf Val. & \bf Test &\bf Non-proj. \%\\\midrule
    \multicolumn{4}{l}{\textit{Low-Resource Test Languages}} \\
    Armenian & IE & Armenian & ArmTDP & 560 & 0 & 470 & 10.2\\
    Breton & IE & Celtic & KEB & 0 & 0 & 888 & 2.7 \\
    Buryat & Mongolic & Mongolic & BDT & 19 & 0 & 908 & 15.6\\
    Faroese & IE & Germanic & OFT & 0 & 0 & 1208 & 2.7 \\
    Kazakh & Turkic & Northwestern & KTB & 31 & 0 & 1047  & 12.1\\
    Upper Sorbian & IE & Slavic & UFAL & 23 & 0 & 623 & 11.3\\
    \midrule
    \multicolumn{4}{l}{\textit{High-Resource Test Languages}} \\
    Finnish & Uralic & Finnic & TDT & 12217 & 1364 & 1555 & 6.3\\
    French & IE & Romance & Spoken & 1153 & 907 & 726& 5.5 \\
    German & IE & Germanic & GSD & 13814 & 799 & 977 & 9.2 \\
    Hungarian & Uralic & Ugric & Szeged & 910 & 441 & 449 & 27.1\\
    Japanese & Japanese & Japanese &  GSD & 7133 & 511 & 551  & 2.6\\
    Persian & IE & Iranian & Seraji & 4798 & 599 & 600 & 6.7\\
    Swedish & IE & Germanic & PUD & 0 & 0 & 1000 & 3.8 \\
    Tamil & Dravidian & Southern & TTB & 400 & 80 & 120 & 2.8\\
    Urdu & IE & Indic & UDTB & 4043 & 552 & 535 & 22.6\\
    Vietnamese & Austro-As. & Viet-Muong & VTB & 1400 & 800 & 800  & 2.9\\
    \midrule
    \multicolumn{4}{l}{\textit{Validation Languages}} \\
    Bulgarian & IE & Slavic & BTB & 8907 & 1115 & 1116 & 3.1\\
    Telugu & Dravidian & South Central &  MTG & 1051 & 131 & 146 & 0.2\\
    \midrule
    \multicolumn{4}{l}{\textit{Train Languages}} \\
    Arabic & Afro-As. & Semitic & PADT & 6075 & 909 & 680  & 9.2\\
    Czech & IE & Slavic & PDT & 68495 & 9270 & 10148 & 11.9\\
    English & IE & Germanic & EWT & 12543 & 2002 & 2077 & 4.8 \\
    Hindi & IE & Indic & HDTB & 13304 & 1659 & 1684& 13.6 \\
    Italian & IE & Romance & ISDT & 13121 & 564 & 482&2.1 \\
    Korean & Korean & Korean & Kaist & 23010 & 2066 & 2287 & 21.7\\
    Norwegian & IE & Germanic & Nynorsk & 14174 & 1890 & 1511  & 8.2\\
    Russian & IE & Slavic & SynTagRus & 48814 & 6584 & 6491 & 8.0\\
    \bottomrule
  \end{tabular}
  \caption{All datasets used during testing (first 16 rows) training and evaluation (final 10 rows), along with the amount of sentences in the dataset and the percentage of non-projective trees throughout that dataset. }
  \label{tab:all_dataset_info}
\end{table*}
We show the full results for each model in \autoref{tab:full20}, \autoref{tab:full40} and \autoref{tab:full80}.

\begin{table*}[!t]
  \centering\small
    \resizebox{\textwidth}{!}{\begin{tabular}{l|ccccccccc}\toprule
\textbf{Language} & \textbf{M.T. only} & \textbf{\textsc{en}} & \textbf{\textsc{hin}} & \textbf{\textsc{ne} (EN)} & \textbf{\textsc{ne} (HIN)} & \textbf{\textsc{maml}} & \textbf{\textsc{maml} (HIN)} & \textbf{\MAMLPrime}
\\ \midrule
 \multicolumn{6}{l}{\it Unseen Languages} \\
Armenian  &   { 4.97}{\scriptsize $\pm$0.007} & { 49.8}{\scriptsize $\pm$0.005} & { 48.41}{\scriptsize $\pm$0.002} & { 63.34}{\scriptsize $\pm$0.002} & { 63.3}{\scriptsize $\pm$0.005} & { \bf 63.84}{\scriptsize $\pm$0.002} & { 63.76}{\scriptsize $\pm$0.003} & { 59.7}{\scriptsize $\pm$0.004} \\
Breton  &   { 10.77}{\scriptsize $\pm$0.019} & { 60.34}{\scriptsize $\pm$0.003} & { 34.06}{\scriptsize $\pm$0.005} & { 61.44}{\scriptsize $\pm$0.005} & { 62.09}{\scriptsize $\pm$0.004} & { \bf 64.18}{\scriptsize $\pm$0.003} & { 61.56}{\scriptsize $\pm$0.002} & { 59.33}{\scriptsize $\pm$0.005} \\
Buryat  &   { 9.63}{\scriptsize $\pm$0.018} & { 23.66}{\scriptsize $\pm$0.002} & { 24.24}{\scriptsize $\pm$0.002} & { 25.56}{\scriptsize $\pm$0.003} & { 25.05}{\scriptsize $\pm$0.003} & { 25.77}{\scriptsize $\pm$0.002} & { \bf 26.27}{\scriptsize $\pm$0.002} & { 26.02}{\scriptsize $\pm$0.004} \\
Faroese  &   { 13.86}{\scriptsize $\pm$0.024} & { 68.5}{\scriptsize $\pm$0.004} & { 50.72}{\scriptsize $\pm$0.004} & { 67.83}{\scriptsize $\pm$0.006} & { 65.31}{\scriptsize $\pm$0.006} & { \bf 68.95}{\scriptsize $\pm$0.003} & { 66.82}{\scriptsize $\pm$0.002} & { 65.3}{\scriptsize $\pm$0.005} \\
Kazakh  &   { 13.97}{\scriptsize $\pm$0.012} & { 47.25}{\scriptsize $\pm$0.004} & { 49.8}{\scriptsize $\pm$0.002} & { 55.02}{\scriptsize $\pm$0.002} & { 53.77}{\scriptsize $\pm$0.003} & { \bf 55.07}{\scriptsize $\pm$0.002} & { 54.23}{\scriptsize $\pm$0.003} & { 53.92}{\scriptsize $\pm$0.005} \\
U.Sorbian  &   { 3.44}{\scriptsize $\pm$0.005} & { 49.29}{\scriptsize $\pm$0.004} & { 36.22}{\scriptsize $\pm$0.003} & { 54.47}{\scriptsize $\pm$0.003} & { 53.36}{\scriptsize $\pm$0.003} & { \bf 56.4}{\scriptsize $\pm$0.004} & { 54.97}{\scriptsize $\pm$0.005} & { 51.67}{\scriptsize $\pm$0.004} \\
Finnish  &   { 6.95}{\scriptsize $\pm$0.014} & { 56.61}{\scriptsize $\pm$0.002} & { 50.49}{\scriptsize $\pm$0.003} & { \bf 64.94}{\scriptsize $\pm$0.003} & { 64.05}{\scriptsize $\pm$0.004} & { 64.89}{\scriptsize $\pm$0.003} & { 64.64}{\scriptsize $\pm$0.002} & { 61.97}{\scriptsize $\pm$0.005} \\
French  &   { 6.81}{\scriptsize $\pm$0.011} & { 65.21}{\scriptsize $\pm$0.001} & { 31.16}{\scriptsize $\pm$0.003} & { 66.55}{\scriptsize $\pm$0.001} & { 64.44}{\scriptsize $\pm$0.002} & { \bf 66.85}{\scriptsize $\pm$0.001} & { 65.73}{\scriptsize $\pm$0.001} & { 63.42}{\scriptsize $\pm$0.003} \\
German  &   { 7.52}{\scriptsize $\pm$0.012} & { 72.47}{\scriptsize $\pm$0.001} & { 44.83}{\scriptsize $\pm$0.004} & { 76.15}{\scriptsize $\pm$0.002} & { 74.4}{\scriptsize $\pm$0.002} & { \bf 76.41}{\scriptsize $\pm$0.002} & { 75.15}{\scriptsize $\pm$0.001} & { 74.38}{\scriptsize $\pm$0.003} \\
Hungarian  &   { 5.58}{\scriptsize $\pm$0.004} & { 56.5}{\scriptsize $\pm$0.003} & { 46.72}{\scriptsize $\pm$0.004} & { \bf 62.93}{\scriptsize $\pm$0.003} & { 60.98}{\scriptsize $\pm$0.002} & { 62.71}{\scriptsize $\pm$0.003} & { 62.51}{\scriptsize $\pm$0.004} & { 58.47}{\scriptsize $\pm$0.002} \\
Japanese  &   { 4.02}{\scriptsize $\pm$0.008} & { 18.87}{\scriptsize $\pm$0.002} & { 40.25}{\scriptsize $\pm$0.005} & { 36.49}{\scriptsize $\pm$0.008} & { 39.97}{\scriptsize $\pm$0.003} & { 39.06}{\scriptsize $\pm$0.003} & { \bf 41.96}{\scriptsize $\pm$0.005} & { 39.72}{\scriptsize $\pm$0.007} \\
Persian  &   { 1.91}{\scriptsize $\pm$0.004} & { 43.42}{\scriptsize $\pm$0.005} & { 28.66}{\scriptsize $\pm$0.004} & { 52.62}{\scriptsize $\pm$0.006} & { \bf 53.78}{\scriptsize $\pm$0.004} & { 52.82}{\scriptsize $\pm$0.005} & { 53.59}{\scriptsize $\pm$0.003} & { 50.31}{\scriptsize $\pm$0.004} \\
Swedish  &   { 5.15}{\scriptsize $\pm$0.008} & { 80.26}{\scriptsize $\pm$0.001} & { 46.96}{\scriptsize $\pm$0.004} & { 80.73}{\scriptsize $\pm$0.001} & { 79.24}{\scriptsize $\pm$0.002} & { \bf 81.36}{\scriptsize $\pm$0.001} & { 79.89}{\scriptsize $\pm$0.001} & { 77.57}{\scriptsize $\pm$0.002} \\
Tamil  &   { 5.18}{\scriptsize $\pm$0.013} & { 31.58}{\scriptsize $\pm$0.005} & { 46.51}{\scriptsize $\pm$0.003} & { 41.12}{\scriptsize $\pm$0.009} & { 39.44}{\scriptsize $\pm$0.006} & { 44.34}{\scriptsize $\pm$0.005} & { 39.57}{\scriptsize $\pm$0.008} & { \bf 46.55}{\scriptsize $\pm$0.01} \\
Urdu  &   { 2.86}{\scriptsize $\pm$0.01} & { 25.71}{\scriptsize $\pm$0.004} & { \bf 67.72}{\scriptsize $\pm$0.001} & { 57.25}{\scriptsize $\pm$0.004} & { 50.64}{\scriptsize $\pm$0.004} & { 55.16}{\scriptsize $\pm$0.004} & { 49.16}{\scriptsize $\pm$0.002} & { 55.4}{\scriptsize $\pm$0.003} \\
Vietnamese  &   { 7.14}{\scriptsize $\pm$0.008} & { 43.24}{\scriptsize $\pm$0.002} & { 26.96}{\scriptsize $\pm$0.002} & { 42.73}{\scriptsize $\pm$0.001} & { 42.13}{\scriptsize $\pm$0.002} & { \bf 43.34}{\scriptsize $\pm$0.001} & { 42.12}{\scriptsize $\pm$0.001} & { 42.62}{\scriptsize $\pm$0.003} \\\midrule
 \multicolumn{6}{l}{\it Validation \& Training Languages} \\
Bulgarian  &   { 8.65}{\scriptsize $\pm$0.01} & { 71.19}{\scriptsize $\pm$0.002} & { 46.76}{\scriptsize $\pm$0.003} & { 78.42}{\scriptsize $\pm$0.003} & { 77.62}{\scriptsize $\pm$0.001} & { \bf 78.64}{\scriptsize $\pm$0.002} & { 78.4}{\scriptsize $\pm$0.001} & { 75.3}{\scriptsize $\pm$0.003} \\
Telugu  &   { 42.36}{\scriptsize $\pm$0.078} & { 64.39}{\scriptsize $\pm$0.018} & { 66.78}{\scriptsize $\pm$0.014} & { 68.5}{\scriptsize $\pm$0.006} & { 64.8}{\scriptsize $\pm$0.008} & { \bf 69.91}{\scriptsize $\pm$0.01} & { 65.8}{\scriptsize $\pm$0.008} & { 67.58}{\scriptsize $\pm$0.008} \\
Arabic  &   { 3.25}{\scriptsize $\pm$0.007} & { 38.53}{\scriptsize $\pm$0.006} & { 20.74}{\scriptsize $\pm$0.004} & { 71.51}{\scriptsize $\pm$0.002} & { 69.76}{\scriptsize $\pm$0.002} & { 68.86}{\scriptsize $\pm$0.002} & { \bf 73.09}{\scriptsize $\pm$0.002} & { 66.4}{\scriptsize $\pm$0.002} \\
Czech  &   { 6.37}{\scriptsize $\pm$0.006} & { 67.3}{\scriptsize $\pm$0.002} & { 43.24}{\scriptsize $\pm$0.002} & { 83.15}{\scriptsize $\pm$0.001} & { 81.65}{\scriptsize $\pm$0.001} & { 82.0}{\scriptsize $\pm$0.001} & { \bf 83.21}{\scriptsize $\pm$0.001} & { 80.06}{\scriptsize $\pm$0.001} \\
English  &   { 8.43}{\scriptsize $\pm$0.008} & { \bf 89.29}{\scriptsize $\pm$0.001} & { 44.48}{\scriptsize $\pm$0.003} & { 82.15}{\scriptsize $\pm$0.004} & { 79.48}{\scriptsize $\pm$0.001} & { 83.74}{\scriptsize $\pm$0.001} & { 81.89}{\scriptsize $\pm$0.001} & { 78.04}{\scriptsize $\pm$0.001} \\
Hindi  &   { 3.38}{\scriptsize $\pm$0.007} & { 35.42}{\scriptsize $\pm$0.002} & { \bf 90.99}{\scriptsize $\pm$0.0} & { 76.56}{\scriptsize $\pm$0.002} & { 74.03}{\scriptsize $\pm$0.004} & { 74.15}{\scriptsize $\pm$0.003} & { 71.33}{\scriptsize $\pm$0.003} & { 74.48}{\scriptsize $\pm$0.004} \\
Italian  &   { 7.15}{\scriptsize $\pm$0.008} & { 82.5}{\scriptsize $\pm$0.001} & { 36.86}{\scriptsize $\pm$0.006} & { \bf 87.34}{\scriptsize $\pm$0.002} & { 85.28}{\scriptsize $\pm$0.001} & { 86.5}{\scriptsize $\pm$0.001} & { 87.18}{\scriptsize $\pm$0.002} & { 83.09}{\scriptsize $\pm$0.002} \\
Korean  &   { 7.82}{\scriptsize $\pm$0.011} & { 36.44}{\scriptsize $\pm$0.004} & { 40.3}{\scriptsize $\pm$0.002} & { 66.35}{\scriptsize $\pm$0.003} & { 68.04}{\scriptsize $\pm$0.003} & { 63.93}{\scriptsize $\pm$0.003} & { \bf 74.08}{\scriptsize $\pm$0.001} & { 63.62}{\scriptsize $\pm$0.005} \\
Norwegian  &   { 5.68}{\scriptsize $\pm$0.013} & { 74.7}{\scriptsize $\pm$0.001} & { 43.7}{\scriptsize $\pm$0.003} & { 80.09}{\scriptsize $\pm$0.001} & { 77.65}{\scriptsize $\pm$0.001} & { 78.67}{\scriptsize $\pm$0.001} & { \bf 81.33}{\scriptsize $\pm$0.002} & { 75.61}{\scriptsize $\pm$0.001} \\
Russian  &   { 6.76}{\scriptsize $\pm$0.013} & { 68.94}{\scriptsize $\pm$0.003} & { 47.29}{\scriptsize $\pm$0.005} & { 80.96}{\scriptsize $\pm$0.001} & { 79.41}{\scriptsize $\pm$0.001} & { 79.93}{\scriptsize $\pm$0.001} & { \bf 81.68}{\scriptsize $\pm$0.001} & { 76.48}{\scriptsize $\pm$0.002} \\
\bottomrule
\end{tabular}}
  \caption{Full meta-testing results for all models and baselines, including validation and training languages, for $|S|=20$. The meta-testing only baseline is denoted as ``M.T. only". }\label{tab:full20}
\end{table*}
\begin{table*}[!t]
  \centering\small
  \resizebox{\textwidth}{!}{\begin{tabular}{l|ccccccccc}\toprule
\textbf{Language} & \textbf{M.T. only} & \textbf{\textsc{en}} & \textbf{\textsc{hin}} & \textbf{\textsc{ne} (EN)} & \textbf{\textsc{ne} (HIN)} & \textbf{\textsc{maml}} & \textbf{\textsc{maml} (HIN)} & \textbf{\MAMLPrime}
\\ \midrule
 \multicolumn{6}{l}{\it Unseen Languages} \\
Armenian  &   { 5.82}{\scriptsize $\pm$0.007} & { 50.59}{\scriptsize $\pm$0.005} & { 48.87}{\scriptsize $\pm$0.002} & { 63.54}{\scriptsize $\pm$0.002} & { 63.41}{\scriptsize $\pm$0.004} & { \bf 64.3}{\scriptsize $\pm$0.002} & { 64.17}{\scriptsize $\pm$0.003} & { 59.85}{\scriptsize $\pm$0.004} \\
Breton  &   { 14.52}{\scriptsize $\pm$0.02} & { 61.32}{\scriptsize $\pm$0.004} & { 36.09}{\scriptsize $\pm$0.005} & { 61.67}{\scriptsize $\pm$0.005} & { 62.4}{\scriptsize $\pm$0.005} & { \bf 65.12}{\scriptsize $\pm$0.003} & { 62.47}{\scriptsize $\pm$0.004} & { 59.96}{\scriptsize $\pm$0.004} \\
Buryat  &   { 13.36}{\scriptsize $\pm$0.017} & { 23.82}{\scriptsize $\pm$0.002} & { 24.71}{\scriptsize $\pm$0.003} & { 25.67}{\scriptsize $\pm$0.003} & { 25.18}{\scriptsize $\pm$0.003} & { 26.38}{\scriptsize $\pm$0.003} & { \bf 26.79}{\scriptsize $\pm$0.003} & { 26.49}{\scriptsize $\pm$0.004} \\
Faroese  &   { 20.4}{\scriptsize $\pm$0.019} & { 69.56}{\scriptsize $\pm$0.004} & { 52.3}{\scriptsize $\pm$0.005} & { 68.12}{\scriptsize $\pm$0.006} & { 65.57}{\scriptsize $\pm$0.006} & { \bf 69.88}{\scriptsize $\pm$0.004} & { 67.31}{\scriptsize $\pm$0.003} & { 65.95}{\scriptsize $\pm$0.004} \\
Kazakh  &   { 17.11}{\scriptsize $\pm$0.014} & { 47.8}{\scriptsize $\pm$0.004} & { 49.9}{\scriptsize $\pm$0.003} & { 55.08}{\scriptsize $\pm$0.002} & { 53.94}{\scriptsize $\pm$0.003} & { \bf 55.46}{\scriptsize $\pm$0.003} & { 54.45}{\scriptsize $\pm$0.004} & { 54.29}{\scriptsize $\pm$0.005} \\
U.Sorbian  &   { 4.49}{\scriptsize $\pm$0.008} & { 50.55}{\scriptsize $\pm$0.005} & { 37.08}{\scriptsize $\pm$0.003} & { 54.7}{\scriptsize $\pm$0.004} & { 53.58}{\scriptsize $\pm$0.003} & { \bf 57.55}{\scriptsize $\pm$0.004} & { 55.64}{\scriptsize $\pm$0.005} & { 52.09}{\scriptsize $\pm$0.005} \\
Finnish  &   { 9.42}{\scriptsize $\pm$0.011} & { 56.99}{\scriptsize $\pm$0.003} & { 50.93}{\scriptsize $\pm$0.003} & { \bf 65.07}{\scriptsize $\pm$0.003} & { 64.2}{\scriptsize $\pm$0.004} & { 65.4}{\scriptsize $\pm$0.003} & { 65.05}{\scriptsize $\pm$0.003} & { 62.26}{\scriptsize $\pm$0.004} \\
French  &   { 8.41}{\scriptsize $\pm$0.019} & { 65.33}{\scriptsize $\pm$0.002} & { 31.59}{\scriptsize $\pm$0.005} & { 66.59}{\scriptsize $\pm$0.001} & { 64.44}{\scriptsize $\pm$0.002} & { \bf 66.97}{\scriptsize $\pm$0.001} & { 65.68}{\scriptsize $\pm$0.002} & { 63.78}{\scriptsize $\pm$0.003} \\
German  &   { 10.4}{\scriptsize $\pm$0.016} & { 72.6}{\scriptsize $\pm$0.001} & { 45.46}{\scriptsize $\pm$0.004} & { 76.17}{\scriptsize $\pm$0.002} & { 74.41}{\scriptsize $\pm$0.002} & { \bf 76.54}{\scriptsize $\pm$0.002} & { 75.23}{\scriptsize $\pm$0.002} & { 74.53}{\scriptsize $\pm$0.003} \\
Hungarian  &   { 6.8}{\scriptsize $\pm$0.007} & { 56.23}{\scriptsize $\pm$0.003} & { 46.97}{\scriptsize $\pm$0.004} & { \bf 63.09}{\scriptsize $\pm$0.003} & { 61.33}{\scriptsize $\pm$0.002} & { 62.81}{\scriptsize $\pm$0.002} & { 62.89}{\scriptsize $\pm$0.003} & { 58.09}{\scriptsize $\pm$0.003} \\
Japanese  &   { 5.85}{\scriptsize $\pm$0.014} & { 20.05}{\scriptsize $\pm$0.003} & { 43.03}{\scriptsize $\pm$0.006} & { 37.15}{\scriptsize $\pm$0.008} & { 40.56}{\scriptsize $\pm$0.002} & { 42.17}{\scriptsize $\pm$0.004} & { \bf 43.61}{\scriptsize $\pm$0.004} & { 41.51}{\scriptsize $\pm$0.006} \\
Persian  &   { 3.32}{\scriptsize $\pm$0.01} & { 44.54}{\scriptsize $\pm$0.004} & { 29.55}{\scriptsize $\pm$0.004} & { 52.72}{\scriptsize $\pm$0.006} & { \bf 53.85}{\scriptsize $\pm$0.004} & { 53.65}{\scriptsize $\pm$0.005} & { 54.02}{\scriptsize $\pm$0.003} & { 50.83}{\scriptsize $\pm$0.005} \\
Swedish  &   { 8.27}{\scriptsize $\pm$0.008} & { 80.41}{\scriptsize $\pm$0.001} & { 47.73}{\scriptsize $\pm$0.003} & { 80.81}{\scriptsize $\pm$0.001} & { 79.32}{\scriptsize $\pm$0.002} & { \bf 81.53}{\scriptsize $\pm$0.001} & { 80.14}{\scriptsize $\pm$0.002} & { 77.94}{\scriptsize $\pm$0.002} \\
Tamil  &   { 9.37}{\scriptsize $\pm$0.018} & { 32.67}{\scriptsize $\pm$0.004} & { 47.35}{\scriptsize $\pm$0.004} & { 41.72}{\scriptsize $\pm$0.009} & { 39.84}{\scriptsize $\pm$0.004} & { 46.73}{\scriptsize $\pm$0.005} & { 40.84}{\scriptsize $\pm$0.006} & { \bf 48.54}{\scriptsize $\pm$0.008} \\
Urdu  &   { 5.88}{\scriptsize $\pm$0.01} & { 26.89}{\scriptsize $\pm$0.005} & { \bf 67.96}{\scriptsize $\pm$0.002} & { 57.36}{\scriptsize $\pm$0.004} & { 50.93}{\scriptsize $\pm$0.004} & { 56.16}{\scriptsize $\pm$0.004} & { 50.16}{\scriptsize $\pm$0.004} & { 55.84}{\scriptsize $\pm$0.003} \\
Vietnamese  &   { 9.65}{\scriptsize $\pm$0.014} & { 43.65}{\scriptsize $\pm$0.002} & { 27.92}{\scriptsize $\pm$0.002} & { 42.82}{\scriptsize $\pm$0.002} & { 42.23}{\scriptsize $\pm$0.003} & { \bf 43.74}{\scriptsize $\pm$0.001} & { 42.37}{\scriptsize $\pm$0.001} & { 43.23}{\scriptsize $\pm$0.004} \\
\midrule 
 \multicolumn{6}{l}{\it Validation \& Training Languages} \\
Bulgarian  &   { 10.87}{\scriptsize $\pm$0.021} & { 71.21}{\scriptsize $\pm$0.002} & { 47.29}{\scriptsize $\pm$0.004} & { 78.42}{\scriptsize $\pm$0.003} & { 77.62}{\scriptsize $\pm$0.001} & { \bf 78.65}{\scriptsize $\pm$0.002} & { 78.39}{\scriptsize $\pm$0.001} & { 75.44}{\scriptsize $\pm$0.003} \\
Telugu  &   { 49.11}{\scriptsize $\pm$0.068} & { 66.64}{\scriptsize $\pm$0.014} & { 67.7}{\scriptsize $\pm$0.013} & { 68.69}{\scriptsize $\pm$0.006} & { 64.75}{\scriptsize $\pm$0.01} & { \bf 70.75}{\scriptsize $\pm$0.012} & { 66.1}{\scriptsize $\pm$0.009} & { 67.97}{\scriptsize $\pm$0.007} \\
Arabic  &   { 4.83}{\scriptsize $\pm$0.006} & { 41.6}{\scriptsize $\pm$0.012} & { 22.65}{\scriptsize $\pm$0.005} & { 71.53}{\scriptsize $\pm$0.002} & { 69.78}{\scriptsize $\pm$0.002} & { 68.95}{\scriptsize $\pm$0.002} & { \bf 73.01}{\scriptsize $\pm$0.003} & { 66.47}{\scriptsize $\pm$0.002} \\
Czech  &   { 7.95}{\scriptsize $\pm$0.008} & { 67.74}{\scriptsize $\pm$0.003} & { 43.92}{\scriptsize $\pm$0.002} & { 83.15}{\scriptsize $\pm$0.001} & { 81.64}{\scriptsize $\pm$0.001} & { 82.03}{\scriptsize $\pm$0.001} & { \bf 83.19}{\scriptsize $\pm$0.001} & { 80.12}{\scriptsize $\pm$0.001} \\
English  &   { 11.01}{\scriptsize $\pm$0.012} & { \bf 89.3}{\scriptsize $\pm$0.001} & { 45.32}{\scriptsize $\pm$0.004} & { 82.21}{\scriptsize $\pm$0.004} & { 79.49}{\scriptsize $\pm$0.001} & { 83.96}{\scriptsize $\pm$0.002} & { 82.05}{\scriptsize $\pm$0.002} & { 78.07}{\scriptsize $\pm$0.001} \\
Hindi  &   { 7.64}{\scriptsize $\pm$0.017} & { 36.64}{\scriptsize $\pm$0.002} & { \bf 90.99}{\scriptsize $\pm$0.0} & { 76.58}{\scriptsize $\pm$0.002} & { 74.24}{\scriptsize $\pm$0.004} & { 74.28}{\scriptsize $\pm$0.002} & { 72.16}{\scriptsize $\pm$0.004} & { 74.5}{\scriptsize $\pm$0.004} \\
Italian  &   { 9.3}{\scriptsize $\pm$0.014} & { 82.68}{\scriptsize $\pm$0.001} & { 38.61}{\scriptsize $\pm$0.007} & { \bf 87.35}{\scriptsize $\pm$0.002} & { 85.28}{\scriptsize $\pm$0.001} & { 86.51}{\scriptsize $\pm$0.001} & { 87.37}{\scriptsize $\pm$0.003} & { 83.11}{\scriptsize $\pm$0.002} \\
Korean  &   { 10.26}{\scriptsize $\pm$0.014} & { 36.77}{\scriptsize $\pm$0.004} & { 40.74}{\scriptsize $\pm$0.003} & { 66.4}{\scriptsize $\pm$0.003} & { 68.07}{\scriptsize $\pm$0.003} & { 64.23}{\scriptsize $\pm$0.003} & { \bf 74.01}{\scriptsize $\pm$0.002} & { 63.91}{\scriptsize $\pm$0.004} \\
Norwegian  &   { 9.24}{\scriptsize $\pm$0.015} & { 74.7}{\scriptsize $\pm$0.002} & { 44.1}{\scriptsize $\pm$0.006} & { 80.06}{\scriptsize $\pm$0.002} & { 77.53}{\scriptsize $\pm$0.004} & { 78.69}{\scriptsize $\pm$0.001} & { \bf 81.2}{\scriptsize $\pm$0.004} & { 75.64}{\scriptsize $\pm$0.001} \\
Russian  &   { 9.09}{\scriptsize $\pm$0.012} & { 69.3}{\scriptsize $\pm$0.003} & { 48.03}{\scriptsize $\pm$0.005} & { 80.98}{\scriptsize $\pm$0.001} & { 79.43}{\scriptsize $\pm$0.001} & { 80.0}{\scriptsize $\pm$0.001} & { \bf 81.66}{\scriptsize $\pm$0.001} & { 76.57}{\scriptsize $\pm$0.002} \\
\bottomrule
\end{tabular}}
  \caption{Full meta-testing results for all models and baselines, including validation and training languages, for $|S|=40$. The meta-testing only baseline is denoted as ``M.T. only".}\label{tab:full40}
\end{table*}
\begin{table*}[!t]
  \centering\small
  \resizebox{\textwidth}{!}{\begin{tabular}{l|ccccccccc}\toprule
\textbf{Language} & \textbf{M.T. only} & \textbf{\textsc{en}} & \textbf{\textsc{hin}} & \textbf{\textsc{ne} (EN)} & \textbf{\textsc{ne} (HIN)} & \textbf{\textsc{maml}} & \textbf{\textsc{maml} (HIN)} & \textbf{\MAMLPrime}
\\ \midrule
 \multicolumn{6}{l}{\it Unseen Languages} \\
Armenian  &   { 8.19}{\scriptsize $\pm$0.006} & { 51.99}{\scriptsize $\pm$0.005} & { 49.7}{\scriptsize $\pm$0.002} & { 63.79}{\scriptsize $\pm$0.002} & { 63.59}{\scriptsize $\pm$0.004} & { \bf 64.78}{\scriptsize $\pm$0.003} & { 64.76}{\scriptsize $\pm$0.003} & { 60.03}{\scriptsize $\pm$0.003} \\
Breton  &   { 22.54}{\scriptsize $\pm$0.018} & { 62.76}{\scriptsize $\pm$0.004} & { 38.95}{\scriptsize $\pm$0.004} & { 62.2}{\scriptsize $\pm$0.006} & { 63.05}{\scriptsize $\pm$0.004} & { \bf 66.14}{\scriptsize $\pm$0.003} & { 63.75}{\scriptsize $\pm$0.004} & { 60.84}{\scriptsize $\pm$0.004} \\
Buryat  &   { 16.87}{\scriptsize $\pm$0.007} & { 24.17}{\scriptsize $\pm$0.003} & { 25.54}{\scriptsize $\pm$0.003} & { 25.88}{\scriptsize $\pm$0.003} & { 25.4}{\scriptsize $\pm$0.003} & { 27.33}{\scriptsize $\pm$0.003} & { \bf 27.37}{\scriptsize $\pm$0.004} & { 27.05}{\scriptsize $\pm$0.004} \\
Faroese  &   { 27.76}{\scriptsize $\pm$0.019} & { 70.59}{\scriptsize $\pm$0.004} & { 54.64}{\scriptsize $\pm$0.005} & { 68.62}{\scriptsize $\pm$0.006} & { 66.17}{\scriptsize $\pm$0.005} & { \bf 71.12}{\scriptsize $\pm$0.004} & { 68.25}{\scriptsize $\pm$0.003} & { 66.79}{\scriptsize $\pm$0.004} \\
Kazakh  &   { 21.89}{\scriptsize $\pm$0.009} & { 49.08}{\scriptsize $\pm$0.004} & { 50.49}{\scriptsize $\pm$0.003} & { 55.23}{\scriptsize $\pm$0.003} & { 54.08}{\scriptsize $\pm$0.003} & { \bf 56.15}{\scriptsize $\pm$0.003} & { 55.0}{\scriptsize $\pm$0.004} & { 54.99}{\scriptsize $\pm$0.005} \\
U.Sorbian  &   { 7.49}{\scriptsize $\pm$0.01} & { 52.11}{\scriptsize $\pm$0.005} & { 38.22}{\scriptsize $\pm$0.004} & { 55.08}{\scriptsize $\pm$0.004} & { 53.94}{\scriptsize $\pm$0.004} & { \bf 58.78}{\scriptsize $\pm$0.005} & { 56.56}{\scriptsize $\pm$0.006} & { 52.38}{\scriptsize $\pm$0.005} \\
Finnish  &   { 11.91}{\scriptsize $\pm$0.012} & { 57.73}{\scriptsize $\pm$0.004} & { 51.79}{\scriptsize $\pm$0.002} & { \bf 65.18}{\scriptsize $\pm$0.003} & { 64.4}{\scriptsize $\pm$0.004} & { 65.82}{\scriptsize $\pm$0.005} & { 65.61}{\scriptsize $\pm$0.004} & { 62.47}{\scriptsize $\pm$0.004} \\
French  &   { 12.42}{\scriptsize $\pm$0.026} & { 65.63}{\scriptsize $\pm$0.002} & { 33.39}{\scriptsize $\pm$0.006} & { 66.65}{\scriptsize $\pm$0.001} & { 64.42}{\scriptsize $\pm$0.002} & { \bf 67.25}{\scriptsize $\pm$0.002} & { 65.69}{\scriptsize $\pm$0.003} & { 64.15}{\scriptsize $\pm$0.003} \\
German  &   { 16.57}{\scriptsize $\pm$0.017} & { 72.93}{\scriptsize $\pm$0.002} & { 46.65}{\scriptsize $\pm$0.003} & { 76.21}{\scriptsize $\pm$0.002} & { 74.46}{\scriptsize $\pm$0.002} & { \bf 76.72}{\scriptsize $\pm$0.002} & { 75.31}{\scriptsize $\pm$0.002} & { 74.72}{\scriptsize $\pm$0.003} \\
Hungarian  &   { 13.0}{\scriptsize $\pm$0.013} & { 56.73}{\scriptsize $\pm$0.003} & { 47.91}{\scriptsize $\pm$0.003} & { \bf 63.21}{\scriptsize $\pm$0.003} & { 61.68}{\scriptsize $\pm$0.002} & { 62.52}{\scriptsize $\pm$0.002} & { 62.91}{\scriptsize $\pm$0.002} & { 57.48}{\scriptsize $\pm$0.004} \\
Japanese  &   { 14.38}{\scriptsize $\pm$0.015} & { 22.8}{\scriptsize $\pm$0.004} & { 46.87}{\scriptsize $\pm$0.004} & { 38.4}{\scriptsize $\pm$0.007} & { 41.58}{\scriptsize $\pm$0.003} & { 46.81}{\scriptsize $\pm$0.003} & { \bf 45.9}{\scriptsize $\pm$0.004} & { 43.87}{\scriptsize $\pm$0.005} \\
Persian  &   { 6.16}{\scriptsize $\pm$0.019} & { 46.4}{\scriptsize $\pm$0.006} & { 31.11}{\scriptsize $\pm$0.01} & { 53.08}{\scriptsize $\pm$0.006} & { \bf 54.01}{\scriptsize $\pm$0.004} & { 54.73}{\scriptsize $\pm$0.006} & { 54.54}{\scriptsize $\pm$0.005} & { 51.07}{\scriptsize $\pm$0.004} \\
Swedish  &   { 12.99}{\scriptsize $\pm$0.011} & { 80.57}{\scriptsize $\pm$0.001} & { 49.15}{\scriptsize $\pm$0.002} & { 80.79}{\scriptsize $\pm$0.002} & { 79.31}{\scriptsize $\pm$0.002} & { \bf 81.59}{\scriptsize $\pm$0.001} & { 80.21}{\scriptsize $\pm$0.002} & { 78.1}{\scriptsize $\pm$0.002} \\
Tamil  &   { 18.46}{\scriptsize $\pm$0.011} & { 34.81}{\scriptsize $\pm$0.007} & { 48.55}{\scriptsize $\pm$0.002} & { 42.88}{\scriptsize $\pm$0.008} & { 40.73}{\scriptsize $\pm$0.004} & { 50.68}{\scriptsize $\pm$0.003} & { 42.81}{\scriptsize $\pm$0.006} & { \bf 50.54}{\scriptsize $\pm$0.008} \\
Urdu  &   { 13.06}{\scriptsize $\pm$0.01} & { 29.3}{\scriptsize $\pm$0.004} & { \bf 68.17}{\scriptsize $\pm$0.004} & { 57.63}{\scriptsize $\pm$0.004} & { 51.5}{\scriptsize $\pm$0.004} & { 57.6}{\scriptsize $\pm$0.004} & { 51.57}{\scriptsize $\pm$0.004} & { 56.28}{\scriptsize $\pm$0.004} \\
Vietnamese  &   { 15.36}{\scriptsize $\pm$0.015} & { 44.28}{\scriptsize $\pm$0.002} & { 29.61}{\scriptsize $\pm$0.002} & { 42.99}{\scriptsize $\pm$0.002} & { 42.46}{\scriptsize $\pm$0.003} & { \bf 44.33}{\scriptsize $\pm$0.002} & { 42.88}{\scriptsize $\pm$0.002} & { 43.78}{\scriptsize $\pm$0.004} \\\midrule 
 \multicolumn{6}{l}{\it Validation \& Training Languages} \\
Bulgarian  &   { 16.26}{\scriptsize $\pm$0.025} & { 71.42}{\scriptsize $\pm$0.003} & { 48.07}{\scriptsize $\pm$0.006} & { 78.43}{\scriptsize $\pm$0.003} & { 77.67}{\scriptsize $\pm$0.002} & { \bf 78.67}{\scriptsize $\pm$0.003} & { 78.45}{\scriptsize $\pm$0.002} & { 75.68}{\scriptsize $\pm$0.003} \\
Telugu  &   { 54.48}{\scriptsize $\pm$0.016} & { 69.08}{\scriptsize $\pm$0.011} & { 68.79}{\scriptsize $\pm$0.01} & { 68.97}{\scriptsize $\pm$0.006} & { 65.05}{\scriptsize $\pm$0.009} & { \bf 71.52}{\scriptsize $\pm$0.012} & { 66.86}{\scriptsize $\pm$0.008} & { 68.41}{\scriptsize $\pm$0.008} \\
Arabic  &   { 9.87}{\scriptsize $\pm$0.015} & { 46.24}{\scriptsize $\pm$0.015} & { 25.5}{\scriptsize $\pm$0.006} & { 71.54}{\scriptsize $\pm$0.002} & { 69.79}{\scriptsize $\pm$0.002} & { 69.07}{\scriptsize $\pm$0.002} & { \bf 73.04}{\scriptsize $\pm$0.002} & { 66.51}{\scriptsize $\pm$0.002} \\
Czech  &   { 10.74}{\scriptsize $\pm$0.012} & { 68.4}{\scriptsize $\pm$0.003} & { 45.28}{\scriptsize $\pm$0.002} & { 83.16}{\scriptsize $\pm$0.001} & { 81.65}{\scriptsize $\pm$0.001} & { 82.04}{\scriptsize $\pm$0.001} & { \bf 83.2}{\scriptsize $\pm$0.001} & { 80.15}{\scriptsize $\pm$0.001} \\
English  &   { 16.86}{\scriptsize $\pm$0.016} & { \bf 89.3}{\scriptsize $\pm$0.001} & { 46.87}{\scriptsize $\pm$0.002} & { 82.32}{\scriptsize $\pm$0.003} & { 79.51}{\scriptsize $\pm$0.001} & { 84.28}{\scriptsize $\pm$0.002} & { 82.08}{\scriptsize $\pm$0.002} & { 78.07}{\scriptsize $\pm$0.002} \\
Hindi  &   { 16.7}{\scriptsize $\pm$0.018} & { 39.25}{\scriptsize $\pm$0.003} & { \bf 90.96}{\scriptsize $\pm$0.0} & { 76.61}{\scriptsize $\pm$0.002} & { 74.65}{\scriptsize $\pm$0.004} & { 74.46}{\scriptsize $\pm$0.003} & { 73.3}{\scriptsize $\pm$0.003} & { 74.63}{\scriptsize $\pm$0.004} \\
Italian  &   { 16.86}{\scriptsize $\pm$0.027} & { 82.96}{\scriptsize $\pm$0.001} & { 41.8}{\scriptsize $\pm$0.009} & { \bf 87.35}{\scriptsize $\pm$0.002} & { 85.29}{\scriptsize $\pm$0.001} & { 86.57}{\scriptsize $\pm$0.002} & { 87.39}{\scriptsize $\pm$0.003} & { 83.17}{\scriptsize $\pm$0.002} \\
Korean  &   { 15.16}{\scriptsize $\pm$0.017} & { 37.77}{\scriptsize $\pm$0.005} & { 41.53}{\scriptsize $\pm$0.003} & { 66.46}{\scriptsize $\pm$0.003} & { 68.16}{\scriptsize $\pm$0.003} & { 64.36}{\scriptsize $\pm$0.004} & { \bf 74.05}{\scriptsize $\pm$0.002} & { 64.21}{\scriptsize $\pm$0.005} \\
Norwegian  &   { 13.08}{\scriptsize $\pm$0.012} & { 74.93}{\scriptsize $\pm$0.002} & { 45.3}{\scriptsize $\pm$0.004} & { 80.08}{\scriptsize $\pm$0.002} & { 77.56}{\scriptsize $\pm$0.004} & { 78.76}{\scriptsize $\pm$0.001} & { \bf 81.22}{\scriptsize $\pm$0.004} & { 75.69}{\scriptsize $\pm$0.001} \\
Russian  &   { 13.37}{\scriptsize $\pm$0.012} & { 69.79}{\scriptsize $\pm$0.003} & { 49.02}{\scriptsize $\pm$0.004} & { 81.01}{\scriptsize $\pm$0.001} & { 79.45}{\scriptsize $\pm$0.001} & { 80.04}{\scriptsize $\pm$0.001} & { \bf 81.67}{\scriptsize $\pm$0.001} & { 76.56}{\scriptsize $\pm$0.002} \\
\bottomrule
\end{tabular}
}
  \caption{Full meta-testing results for all models and baselines, including validation and training languages, for $|S|=80$. The meta-testing only baseline is denoted as ``M.T. only".}\label{tab:full80}
\end{table*}
\clearpage\newpage

\clearpage
\section{Results for Italian/Czech pre-training}
\label{sec:czech}
We repeat the setup that is discussed in the main paper on another pair of pre-training languages: Italian and Czech. These two languages are, as are English and Hindi, Indo-European, but vary in amount of non-projective sentences within their UD treebanks: 2.1\% sentences are non-projective for the Italian UD dataset used, and 11.9\% for the Czech dataset (see also \autoref{tab:all_dataset_info}). This allows us to further corroborate our findings on non-projectivity.

We randomly take 13 thousand sentences from the Czech training set to match the size of the other three pre-training sets used and verify that the percentage of non-projective sentences is of the same magnitude on this new training set. We run a separate, smaller hyperparameter search for these experiments. All hyperparameters for the monolingual (\textsc{cz, it}), non-episodic (\textsc{ne}), and meta-learning (\textsc{maml}) models are selected using meta-validation. These hyperparameters can be seen in \autoref{tab:finals_czech}. 

\begin{table}[!h]
  \centering\small
  \begin{tabular}{l|rr|rr}
  \toprule
    & \multicolumn{2}{c}{\bf Inner/Test LR} & \multicolumn{2}{c}{\bf Outer LR}\\
    & Decoder & BERT & Decoder & BERT \\
    \midrule
    \textsc{it}/\textsc{cz} & 1e-4  &1e-4  & n/a & n/a  \\
    \textsc{ne} (it/cz) & 5e-4 & 1e-4 & 1e-4 & 7e-6 \\
    \textsc{maml} (it/cz) & 1e-3 & 5e-4 & 5e-4 & 1e-5 \\
    \bottomrule
  \end{tabular}
  \caption{Final hyperparameters in the Italian/Czech setup, as selected by few-shot performance on the meta-validation set. Inner loop/Test learning rates are used with SGD, outer loop LRs are used with the Adam optimizer.}
  \label{tab:finals_czech}
\end{table}

\subsection{Performance}
The full results can be seen in \autoref{tab:full_italian} and \autoref{tab:full_czech}. The results are similar to those in \autoref{tab:testresults} for both the low-resource and the high-resource category. \textsc{maml} slightly outperforms the corresponding non-episodic baseline \textsc{ne}, especially on unrelated languages from Italian and Czech, such as Japanese.

\subsection{Projectivity}
For \textsc{maml} with Italian pre-training, Spearman's $\rho=0.43$ ($p=0.028$). 
For \textsc{maml} with Czech pre-training, the effect is not significant $\rho=0.3$ ($p=0.1349$).

These correlations were, as in the original experiments, calculated using the training language set as well as the testing language set. Excluding training languages in this calculation, the correlation is weaker for Italian pre-training $\rho=0.39,p=0.048$ and non-existent for Czech pre-training ($\rho=0.03$). This
again suggests that a model trained on a mostly projective language can benefit more from further training on non-projective languages than vice versa.

\clearpage
\begin{table*}[!t]
  \centering
  \small
  \begin{tabular}{l|rrr|rrr|rrr}\toprule
  & \multicolumn{3}{c|}{$|S|=20$} &  \multicolumn{3}{c|}{$|S|=40$} & \multicolumn{3}{c}{$|S|=80$} \\
  \textbf{Language} & \textbf{\textsc{it}} & \textbf{\textsc{ne} } & \textbf{\textsc{maml} } & \textbf{\textsc{it}} & \textbf{\textsc{ne} } & \textbf{\textsc{maml} } & \textbf{\textsc{it}} & \textbf{\textsc{ne} } & \textbf{\textsc{maml} }\\\midrule
    \multicolumn{5}{l}{ \it Low-Resource Languages } &\\  
Armenian  &   { 53.06} & {\bf 64.61} & { 63.7} & { 53.47} & {\bf 64.97} & { 64.28} & { 54.51} & {\bf 65.42} & { 64.84} \\
Breton  &   { 57.75} & {61.36} & {\bf 64.13} & { 59.27} & { 62.75} & {\bf 65.95} & { 61.95} & { 64.63} & {\bf 67.48} \\
Buryat  &   { 23.81} & { 26.7} & {\bf 27.43} & { 24.46} & { 27.27} & {\bf 29.27} & { 25.08} & { 28.22} & \underline{\bf 30.86} \\
Faroese  &   { 65.26} & { 69.2} & {\bf 69.39} & { 66.67} & { 69.85} & \underline{\bf 70.94} & { 68.22} & { 70.64} & \underline{\bf 72.36} \\
Kazakh  &   { 45.69} & {\bf 55.44} & { 55.35} & { 46.36} & { 56.08} & {\bf 56.6} & { 47.71} & { 57.01} & {\bf 58.49} \\
U.Sorbian &   { 50.61} & { 55.46} & \underline{\bf 56.79} & { 51.97} & { 56.35} & \underline{\bf 59.2} & { 53.25} & { 57.79} & \underline{\bf 62.3} \\
  \it Mean & 49.36 & 55.46 & 56.13 & 50.37 & 56.21 & 57.71 & 51.79 & 57.28 & 59.39  \\  \midrule
  \multicolumn{5}{l}{ \it High-Resource Unseen Languages } &\\
Finnish  &   { 58.98} & {\bf 66.77} & { 66.44} & { 59.31} & {\bf 67.26} & { 67.09} & { 59.81} & {\bf 67.57} & { 67.44} \\
French  &   { 64.12} & {\bf 66.65} & { 65.99} & { 64.97} & {\bf 66.69} & { 66.26} & { 65.87} & { 66.63} & {\bf 67.17} \\
German  &   { 74.3} & { 75.69} & {\bf 76.18} & { 74.29} & { 75.83} & {\bf 76.55} & { 74.34} & { 76.03} & {\bf 76.9} \\
Hungarian  &   { 58.21} & {\bf 62.97} & { 62.87} & { 58.22} & {\bf 63.43} & { 61.94} & { 58.54} & {\bf 63.45} & { 60.34} \\
Japanese  &   { 15.2} & { 40.7} & \underline{\bf 46.71} & { 16.35} & { 44.07} & \underline{\bf 54.38} & { 18.53} & { 48.88} & \underline{\bf 60.49} \\
Persian  &   { 46.37} & { 53.67} & {\bf 54.81} & { 47.0} & { 54.35} & {\bf 55.96} & { 48.25} & { 55.49} & {\bf 57.93} \\
Swedish  &   { 75.98} & {\bf 80.85} & { 80.56} & { 76.22} & {\bf 80.98} & { 80.95} & { 76.31} & { 81.09} & {\bf 81.29} \\
Tamil  &   { 28.86} & { 44.84} & {\bf 48.1} & { 30.88} & { 47.47} & {\bf 53.43} & { 34.91} & { 50.58} & {\bf 56.39} \\
Urdu  &   { 19.81} & { 57.05} & {\bf 57.21} & { 20.6} & { 57.92} & {\bf 59.03} & { 22.29} & { 58.95} & {\bf 60.69} \\
Vietnamese  &   { 42.7} & { 42.96} & {\bf 43.94} & { 42.95} & { 43.48} & {\bf 44.99} & { 43.64} & { 44.31} & \underline{\bf 46.59} \\
  \it Mean & 48.45 & 59.21 & 60.28 & 49.08 & 60.15 & 62.06 & 50.25 & 61.3 & 63.52\\ 
  \midrule
 \it Mean & 48.79 & 57.82 & 58.73 & 49.56 & 58.67 & 60.43 & 50.83 & 59.79 & 61.97 \\ \midrule
  \multicolumn{5}{l}{ \it Validation \& Training Languages } &\\
Bulgarian  &   { 76.26} & {\bf 78.49} & { 78.04} & { 76.32} & {\bf 78.53} & { 78.36} & { 76.48} & { 78.59} & {\bf 78.7} \\
Telugu  &   { 62.07} & { 69.0} & {\bf 71.46} & { 64.09} & { 69.51} & {\bf 71.37} & { 65.8} & { 70.74} & {\bf 73.27} \\
Arabic  &   { 45.03} & {\bf 71.88} & { 69.68} & { 48.73} & {\bf 71.97} & { 69.94} & { 53.5} & {\bf 72.01} & { 70.19} \\
Czech  &   { 71.82} & {\bf 81.14} & { 80.97} & { 71.93} & {\bf 81.15} & { 81.03} & { 72.28} & {\bf 81.2} & { 81.12} \\
English  &   { 72.04} & {\bf 81.36} & { 81.26} & { 72.37} & {\bf 81.36} & { 81.32} & { 72.88} & {\bf 81.39} & { 81.28} \\
Hindi  &   { 29.36} & {\bf 75.98} & { 74.24} & { 30.45} & {\bf 76.06} & { 74.63} & { 32.49} & {\bf 76.16} & { 75.04} \\
Italian  &   {\bf 93.32} & { 90.61} & { 91.38} & {\bf 93.32} & { 90.7} & { 91.55} & {\bf 93.32} & { 90.8} & { 91.75} \\
Korean  &   { 32.45} & {\bf 64.79} & { 63.62} & { 32.86} & {\bf 65.1} & { 64.1} & { 34.1} & {\bf 65.3} & { 64.37} \\
Norwegian  &   { 74.53} & {\bf 79.28} & { 78.35} & { 74.67} & {\bf 79.33} & { 78.61} & { 74.91} & {\bf 79.47} & { 78.79} \\
Russian  &   { 71.78} & {\bf 80.9} & { 80.06} & { 72.05} & {\bf 80.96} & { 80.24} & { 72.41} & {\bf 81.01} & { 80.32} \\
  \bottomrule
\end{tabular}
  \caption{ Results for Italian pre-training. Mean LAS aligned accuracy per support set size $|S|$ for all languages. Best results per category are bolded. Significant results are underlined ($p<0.005$).}
  \label{tab:full_italian}
\end{table*}
\begin{table*}[!t]
  \centering
  \small
 
\begin{tabular}{l|rrr|rrr|rrr}\toprule
  & \multicolumn{3}{c|}{$|S|=20$} & \multicolumn{3}{c}{$|S|=40$} & \multicolumn{3}{c}{$|S|=80$} \\
  \textbf{Language} & \textbf{\textsc{cz}} & \textbf{\textsc{ne} } & \textbf{\textsc{maml} } & \textbf{\textsc{cz}} & \textbf{\textsc{ne} } & \textbf{\textsc{maml} } & \textbf{\textsc{cz}} & \textbf{\textsc{ne} } & \textbf{\textsc{maml} } \\\midrule
  \multicolumn{5}{l}{ \it Low-Resource Languages } &\\
 Armenian  &   { 55.98} & { \bf 65.01} & { 64.12} & { 56.75} & { \bf 65.26} & { 64.91} & { 58.04} & { \bf 65.71} & { 65.66} \\
Breton  &   { 52.42} & { 62.44} & { \bf 64.48} & { 55.09} & { 63.61} & { \bf 65.49} & { 58.23} & { 64.7} & { \bf 67.02} \\
Buryat  &   { 23.15} & { 27.54} & { \bf 27.78} & { 23.71} & { 27.81} & \underline{ \bf 29.44} & { 24.42} & { 28.86} & \underline{ \bf 31.04} \\
Faroese  &   { 59.19} & { 67.65} & \underline{ \bf 68.24} & { 60.38} & { 68.89} & { \bf 70.33} & { 61.65} & { 69.97} & \underline{ \bf 71.73} \\
Kazakh  &   { 44.61} & { \bf 55.7} & { 54.58} & { 45.46} & { \bf 55.84} & { 55.49} & { 46.87} & { 56.77} & { \bf 57.24} \\
U.Sorbian &   { 56.79} & { 57.0} & { \bf 58.18} & { 57.94} & { 57.92} & { \bf 60.46} & { 59.53} & { 59.12} & \underline{ \bf 63.68} \\
  \it Mean &48.69 & 55.89 & 56.23 & 49.89 & 56.56 & 57.69 & 51.46 & 57.52 & 59.39 \\
  \midrule
  \multicolumn{5}{l}{ \it High-Resource Unseen Languages } &\\ 
  Finnish  &   { 56.11} & { \bf 66.15} & { 65.89} & { 56.46} & { \bf 66.73} & { 66.57} & { 56.95} & { 66.84} & { \bf 67.0} \\
French  &   { 54.69} & { \bf 65.67} & { 64.5} & { 55.66} & { \bf 65.26} & { 64.63} & { 57.63} & { \bf 65.37} & { 65.17} \\
German  &   { 65.38} & { 75.3} & { \bf 75.71} & { 66.05} & { 75.48} & { \bf 75.97} & { 67.56} & { 75.69} & { \bf 76.42} \\
Hungarian  &   { 52.11} & { \bf 62.62} & { 61.11} & { 52.61} & { \bf 62.76} & { 60.99} & { 54.74} & { \bf 63.13} & { 60.3} \\
Japanese  &   { 12.66} & { 40.96} & { \bf 45.06} & { 13.84} & { 43.81} & \underline{ \bf 51.61} & { 16.69} & { 48.47} & \underline{ \bf 57.6} \\
Persian  &   { 50.77} & { 54.23} & { \bf 55.06} & { 51.01} & { 55.14} & { \bf 56.24} & { 51.72} & { 56.06} & { \bf 58.29} \\
Swedish  &   { 67.61} & { \bf 81.04} & { 79.95} & { 67.88} & { \bf 81.61} & { 80.4} & { 68.55} & \underline{ \bf 81.45} & { 80.89} \\
Tamil  &   { 34.53} & { 46.46} & { \bf 49.42} & { 36.54} & { 46.94} & { \bf 54.03} & { 39.08} & { 52.15} & { \bf 56.61} \\
Urdu  &   { 22.26} & { 57.54} & { \bf 57.55} & { 22.98} & { 58.91} & { \bf 59.5} & { 24.44} & { 59.28} & { \bf 60.98} \\
Vietnamese  &   { 39.52} & { 43.14} & \underline{ \bf 44.35} & { 40.03} & { 43.72} & \underline{ \bf 45.0} & { 40.85} & { 44.34} & { \bf 46.24} \\
  \it Mean &  45.56 & 59.31 & 59.86 & 46.31 & 60.04 & 61.49 & 47.82 & 61.28 & 62.95\\ 
  \midrule
  \it Mean & 46.74 & 58.03 & 58.5 & 47.65 & 58.73 & 60.07 & 49.18 & 59.87 & 61.62 \\
  \midrule 
  \multicolumn{5}{l}{ \it Validation \& Training Languages } &\\
  Bulgarian  &   { 76.86} & { \bf 77.92} & { 77.32} & { 76.85} & { \bf 77.95} & { 77.57} & { 76.84} & { \bf 78.06} & { 78.0} \\
Telugu  &   { 61.63} & { 68.21} & { \bf 68.55} & { 63.28} & { 68.88} & { \bf 69.97} & { 65.36} & { 69.45} & { \bf 71.89} \\
Arabic  &   { 57.77} & { \bf 72.3} & { 69.78} & { 58.15} & { \bf 72.55} & { 70.17} & { 58.7} & { \bf 72.31} & { 70.38} \\
Czech  &   { \bf 90.15} & { 85.51} & { 85.95} & { \bf 90.15} & { 86.34} & { 86.29} & { \bf 90.15} & { 85.72} & { 86.65} \\
English  &   { 58.9} & { \bf 80.26} & { 79.75} & { 60.07} & { \bf 80.37} & { 79.89} & { 62.18} & { \bf 80.4} & { 79.97} \\
Hindi  &   { 30.97} & { \bf 76.08} & { 74.09} & { 31.9} & { \bf 76.21} & { 74.36} & { 33.83} & { \bf 76.37} & { 74.79} \\
Italian  &   { 69.39} & { \bf 85.74} & { 85.01} & { 70.79} & { \bf 85.66} & { 85.06} & { 73.52} & { \bf 85.72} & { 85.18} \\
Korean  &   { 32.47} & { \bf 65.4} & { 63.7} & { 32.89} & { \bf 66.29} & { 64.31} & { 33.93} & { \bf 66.15} & { 64.63} \\
Norwegian  &   { 62.08} & { \bf 79.47} & { 77.48} & { 62.58} & { \bf 79.63} & { 77.73} & { 63.49} & { \bf 79.57} & { 77.89} \\
Russian  &   { 76.77} & { \bf 81.65} & { 80.34} & { 76.84} & { \bf 82.17} & { 80.51} & { 76.95} & { \bf 81.72} & { 80.6} \\
  \bottomrule
\end{tabular}
  \caption{Results for Czech pre-training. Mean LAS aligned accuracy per support set size $|S|$ for all languages. Best results per category are bolded. Significant results are underlined ($p<0.005$).}
  \label{tab:full_czech}
\end{table*}

\end{document}